%% file: main.tex
\documentclass[10pt,twocolumn,letterpaper]{article}

\usepackage[pagenumbers]{iccv} 
\usepackage{marvosym}
\input{preamble}

\definecolor{iccvblue}{rgb}{0.21,0.49,0.74}
\usepackage[pagebackref,breaklinks,colorlinks,allcolors=iccvblue]{hyperref}

\begin{document}

\title{\emph{\textcolor{titlegreen}{Any2Caption}}\AnyCapLogo : Interpreting Any Condition to Caption \\ for Controllable Video Generation
}


\author{
Shengqiong Wu$^{1,2}$\thanks{Work done during internship at Kuaishou Technology} \quad 
Weicai Ye$^{1,\textrm{\Letter}}$ \quad 
Jiahao Wang$^1$ \quad 
Quande Liu$^1$ \quad
Xintao Wang$^1$ \quad \\
Pengfei Wan$^{1}$ \quad
Di Zhang$^1$ \quad
Kun Gai$^1$ \quad 
Shuicheng Yan$^{2}$\quad 
Hao Fei$^{2,\textrm{\Letter}}$ \quad 
Tat-Seng Chua$^{2}$ \\
\textnormal{$^1$Kuaishou Technology} \qquad \textnormal{$^2$National University of Singapore} 
\\
\textbf{\small \urlstyle{tt}{\url{https://sqwu.top/Any2Cap/}}}
}

\twocolumn[{%
\renewcommand\twocolumn[1][]{#1}%
\maketitle
\vspace{-8mm}
\begin{center}
   \includegraphics[width=0.99\linewidth]{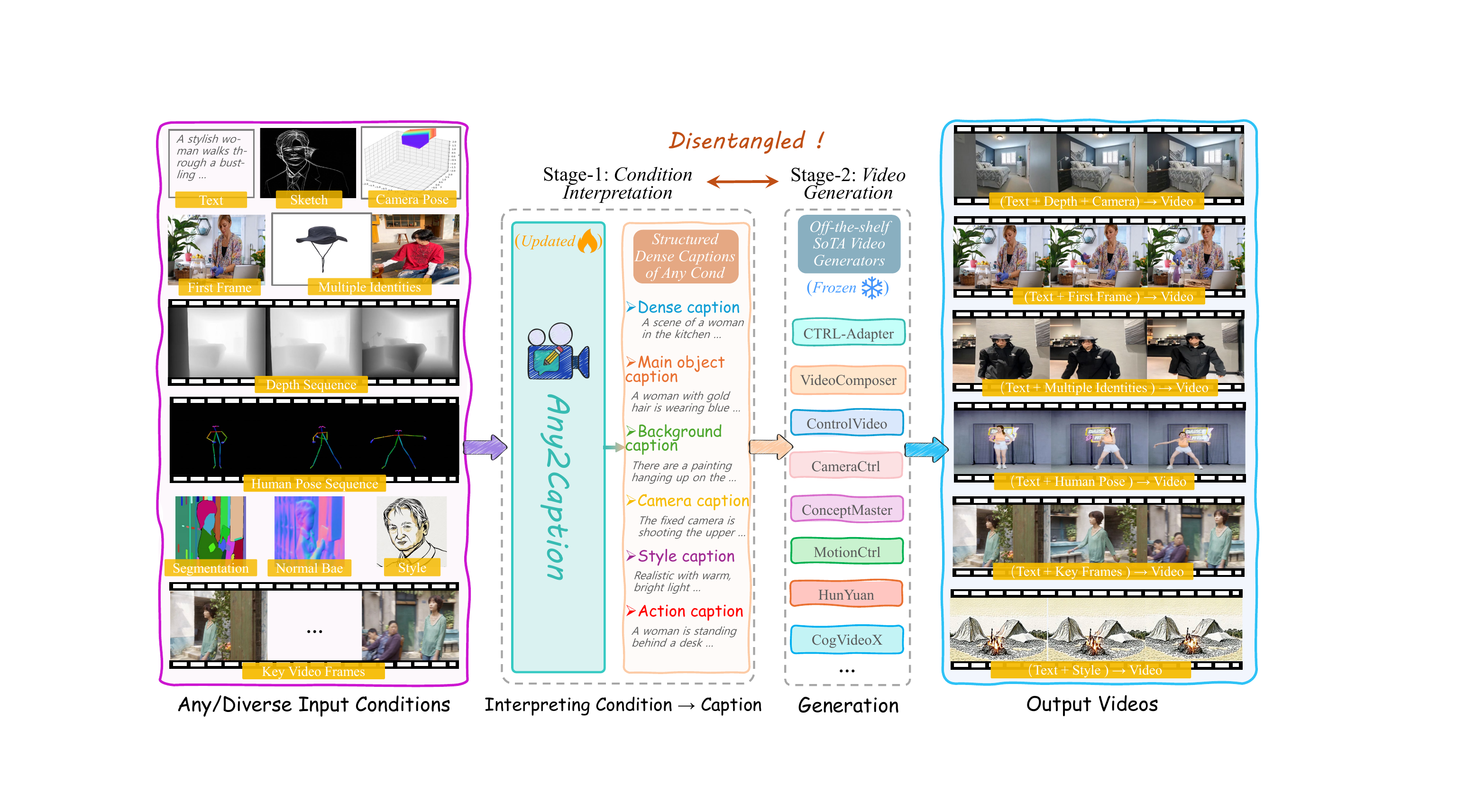}
    \vspace{-2mm}
    \captionof{figure}{We propose \texttt{Any2Caption}, an efficient and versatile framework for interpreting diverse conditions to structured captions, which then can be fed into any video generator to generate highly controllable videos.}
\label{fig:model_teaser}
\end{center}%
}]

\def\thefootnote{$\ast$}\footnotetext{Work done during internship at Kuaishou Technology.}\def\thefootnote{\textrm{\Letter}}\footnotetext{Corresponding Author.}

\begin{abstract}
To address the bottleneck of accurate user intent interpretation within the current video generation community, we present \textbf{\texttt{Any2Caption}}, a novel framework for controllable video generation under any condition.
The key idea is to decouple various condition interpretation steps from the video synthesis step.  
By leveraging modern multimodal large language models (MLLMs), \texttt{Any2Caption} interprets diverse inputs—text, images, videos, and specialized cues such as region, motion, and camera poses—into dense, structured captions that offer backbone video generators with better guidance.  
We also introduce \textbf{\texttt{Any2CapIns}}, a large-scale dataset with 337K instances and 407K conditions for any-condition-to-caption instruction tuning.  
Comprehensive evaluations demonstrate significant improvements of our system in controllability and video quality across various aspects of existing video generation models. 
\end{abstract}

\vspace{-4mm}
\section{Introduction}
\label{sec:intro}

Video serves as a fundamental medium for capturing real-world dynamics, making diverse and controllable video generation a key capability for modern artificial intelligence (AI) systems. 
Recently, video generation has gained significant attention, driven by advancements in Diffusion Transformers (DiT) \cite{Open-Sora,DiVE-abs-2409-01595,Make-A-Video,Latte-abs-2401-03048,kuaishou}, which have demonstrated the ability to generate realistic, long-duration videos from text prompts. 
These advancements have even led to industrial applications, such as filmmaking.
However, we observe that a major bottleneck in the current video generation community lies in \textbf{accurately interpreting user intention}, so as to produce high-quality, controllable videos.

In text-to-video (T2V) generation, studies \cite{CogVideoX,MiraData-Ju00YWZX0S24,CogVideo-Hong0ZL023} have suggested that detailed prompts, specifying objects, actions, attributes, poses, camera movements, and style, significantly enhance both controllability and video quality. 
Thus, a series of works have explored video recaption techniques (e.g., ShareGPT4Video \cite{ShareGPT4Video}, MiraData \cite{MiraData-Ju00YWZX0S24}, and InstanceCap \cite{InstanceCap}) to build dense structured captions for optimizing generative models.
While dense captions are used during training, in real-world inference scenarios, users most likely provide concise or straightforward input prompts \cite{InstanceCap}.
Such a gap inevitably weakens instruction following and leads to suboptimal generation due to an incomplete understanding of user intent.
To combat this, there are two possible solutions,
manual prompt refinement or automatic prompt enrichment \cite{InstanceCap, CogVideoX} using large language models (LLMs).
Yet, these approaches either require substantial human effort or risk introducing noise from incorrect prompt interpretations. 
As a result, this limitation in precisely interpreting user intent hinders the adoption of controllable video generation for demanding applications such as anime creation and filmmaking.

In addition, to achieve more fine-grained controllable video generation, one effective strategy is to provide additional visual conditions besides text prompts—such as reference images \cite{Tune-A-Video-WuGWLGSHSQS23,I2V-Adapter-GuoZHGDWZLHZHM24}, identity \cite{abs-2411-17440,abs-2404-15275,abs-2402-09368}, style \cite{StyleMaster,StyleCrafter-LiuXZCXWWSY24}, human pose \cite{MaHCWC0C24,DreamPose-KarrasHWK23}, or camera \cite{CameraCtrl,CamI2V}—or even combinations of multiple conditions together \cite{ControlVideo,Ctrl-Adapter,VideoComposer-WangYZCWZSZZ23}.
This multimodal conditioning approach aligns well with real-world scenarios, as users quite prefer interactive ways to articulate their creative intent. 
Several studies have examined video generation under various conditions, such as VideoComposer \cite{VideoComposer-WangYZCWZSZZ23}, Ctrl-Adapter \cite{Ctrl-Adapter}, and ControlVideo \cite{ControlVideo}. 
Unfortunately, these methods tend to rely on the internal encoders of Diffusion/DiT to parse rich heterogeneous input conditions with intricate requirements (e.g., multiple object IDs, and complex camera movements).
Before generation, the model must accurately interpret the semantics of varied visual conditions in tandem with textual prompts. 
Yet even state-of-the-art (SoTA) DiT backbones have limited capacity for reasoning across different input modalities, resulting in suboptimal generation quality.

This work is dedicated to addressing these bottlenecks of any-conditioned video generation.
Our core idea is to \textbf{decouple the first job of interpreting various conditions from the second job of video generation}, motivated by two important observations: 
\begin{compactitem}
    \item[a)] SoTA video generation models (e.g., DiT) already excel at producing high-quality videos when presented with sufficiently rich text captions;
    \item[b)] Current MLLMs have demonstrated robust vision-language comprehension. 
\end{compactitem}
Based on these, we propose \textbf{\texttt{Any2Caption}}, an MLLM-based universal condition interpreter designed not only to handle text, image, and video inputs but also equipped with specialized modules for motion and camera pose inputs. 
As illustrated in Fig. \ref{fig:model_teaser}, \texttt{Any2Caption} takes as inputs any/diverse condition (or combination), and produces a densely structured caption, which is then passed on to any backbone video generators for controllable, high-quality video production. 
As \texttt{Any2Caption} disentangles the role of complex interpretation of multimodal inputs from the backbone generator, it advances in seamlessly integrating into a wide range of well-trained video generators without the extra cost of fine-tuning.

To facilitate the any-to-caption instruction tuning for \texttt{Any2Caption}, we construct \textbf{\texttt{Any2CapIns}}, a large-scale dataset that converts a concise user prompt and diverse non-text conditions into detailed, structured captions. 
Concretely, the dataset encompasses four main categories of conditions: depth maps, multiple identities, human poses, and camera poses. 
Through extensive manual labeling combined with automated annotation by GPT-4V \cite{Gpt-4v}, followed by rigorous human verification, we curate a total of \textbf{337K} high-quality instances, with \textbf{407K} condition annotations, with the short prompts and structured captions averaging 55 and 231 words, respectively.
In addition, we devise a comprehensive evaluation strategy to thoroughly measure the model's capacity for interpreting user intent under these various conditions.

\begin{figure*}[!t]
\centering
\includegraphics[width=0.99\textwidth]{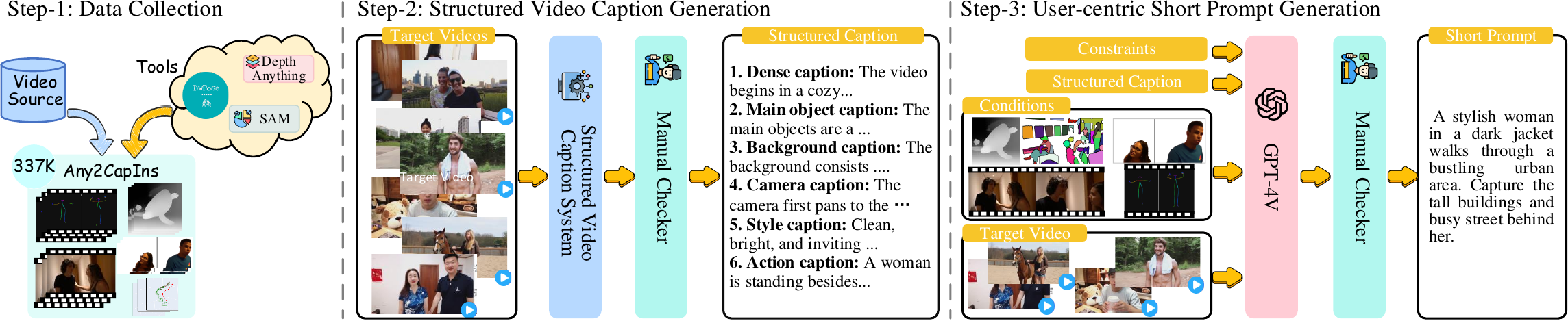}
\vspace{-2mm}
\caption{
The pipeline for constructing the \texttt{Any2CapIns} dataset involves three key steps: 1) data collection, 2) structured video caption generation, and 3) user-centric short prompt generation.
}
\vspace{-3mm}
\label{fig:Any2CapIns}
\end{figure*}

Experimentally, we first validate \texttt{Any2Caption} on our \texttt{Any2CapIns}, where results demonstrate that it achieves an impressive captioning quality that can faithfully reflect the original input conditions. 
We then experiment with integrating \texttt{Any2Caption} with multiple SoTA video generators, finding that 
(a) the long-form semantically rich prompts produced by \texttt{Any2Caption} are pivotal for generating high-quality videos under arbitrary conditions, 
and (b) \texttt{Any2Caption} consistently enhances performance across different backbone models, yielding noticeably improved outputs. 
Furthermore, \texttt{Any2Caption} shows a pronounced advantage when handling multiple combined conditions, effectively interpreting and synthesizing intricate user constraints into captions that align closely with user expectations.
Our contributions are threefold:
\begin{compactitem}
    \item We for the first time pioneer a novel \emph{any-condition-to-caption} paradigm of video generation, which bridges the gap between user-provided multimodal conditions and structured video generation instructions, leading to highly-controllable video generation.
    
    \item We propose \texttt{Any2Caption} to effectively integrate and comprehend diverse multimodal conditions, producing semantically enriched, long-form, structured captions, which consistently improve both condition flexibility and video quality.
    \texttt{Any2Caption} can also be widely integrated as a plug-in module for any existing video generation framework.

    \item We introduce \texttt{Any2CapIns}, a large-scale, high-quality benchmark dataset for the any-condition-to-caption task, and also establish a suite of evaluation metrics to rigorously assess the quality and fidelity of condition-based caption generation.
\end{compactitem}

\vspace{-1mm}
\section{Related Work}
\label{sec:related_work}
\vspace{-1mm}

Controllable video generation \cite{abs-2411-04928,abs-2408-17424,abs-2411-18281,CameraCtrl} has long been a central topic in AI.  
Recent advanced DiT methods, such as OpenAI's Sora \cite{sora} and HunyuanVideo \cite{kong2024hunyuanvideo}, yield photorealistic videos over extended durations.  
Early work focused on text-controlled video generation \cite{Make-A-Video,CogVideo-Hong0ZL023}, the prevalent approach.  
Yet, text prompts alone may insufficiently capture user intent, spurring exploration of additional inputs including static images \cite{Tune-A-Video-WuGWLGSHSQS23,I2V-Adapter-GuoZHGDWZLHZHM24}, sketches \cite{ControlVideo,VideoComposer-WangYZCWZSZZ23}, human poses \cite{PoseCrafter-ZhongZYYZL24,MaHCWC0C24,DreamPose-KarrasHWK23}, camera views \cite{CamI2V,CameraCtrl}, and even extra videos \cite{KaraKYRY24,DengWZTT24,I2VGen-XL}.  
Thus, unifying these diverse conditions into an ``any-condition'' framework is highly valuable.

Recent works such as VideoComposer \cite{VideoComposer-WangYZCWZSZZ23}, Ctrl-Adapter \cite{Ctrl-Adapter}, and ControlVideo \cite{ControlVideo} have explored any-condition video generation.  
However, they face challenges in controlling multiple modalities due to the limited interpretability of text encoders in Diffusion or DiT.  
Motivated by existing MLLMs' multimodal reasoning \cite{llava-LiuLLL24,Video-LLaVA-LinYZCNJ024,Qwen2-VL}, we propose leveraging an MLLM to consolidate all possible conditions into structured dense captions for better controllable generation.  
SoTA DiT models already exhibit the capacity to interpret dense textual descriptions as long as the input captions are sufficiently detailed in depicting both the scene and the intended generation goals. 
Thus, our MLLM-based encoder alleviates the comprehension bottleneck, enabling higher-quality video generation.  
To our knowledge, this is the first attempt in the field of any-condition video generation.  
Moreover, as the captioning stage is decoupled from backbone DiT, \texttt{Any2Caption} can integrate with existing video generation solutions without additional retraining.

Our approach also relates to video recaptioning, as our system produces dense captions from given conditions.  
In text-to-video settings, prior work \cite{InstanceCap,CogVideoX,abs-2407-02371,IslamHYNTB24} shows that recaptioning yields detailed annotations that improve DiT training.  
For instance, ShareGPT4Video \cite{ShareGPT4Video} uses GPT-4V \cite{Gpt-4v} to reinterpret video content, while MiraData \cite{MiraData-Ju00YWZX0S24} and InstanceCap \cite{InstanceCap} focus on structured and instance-consistent recaptioning.  
Unlike these methods, we avoid retraining powerful DiT models with dense captions by training an MLLM as an any-condition encoder on pairs of short, dense captions that are easier to obtain.  
Moreover, recaptioning entire videos can introduce noise or hallucinations that undermine DiT training, whereas our framework sidesteps this risk.  
Finally, while previous studies rely on dense-caption-trained DiT models, the real-world user concise prompts might create a mismatch that degrades generation quality.

\vspace{-1mm}
\section{Any2CapIns Dataset Construction}
\label{sec:dataset}
\vspace{-1mm}

While relevant studies recaption target videos for dense captions for enhanced T2V generation \cite{ShareGPT4Video,InstanceCap,MiraData-Ju00YWZX0S24}, these datasets exhibit two key limitations: 
(1) the absence of non-text conditions, and (2) short prompts that do not account for interactions among non-text conditions, potentially leading to discrepancies in real-world applications.
To address these limitations, we introduce a new dataset, \texttt{Any2CapIns}, specifically designed to incorporate diverse multimodal conditions for generating structured video captions. 
The dataset is constructed through a three-step process (cf. Fig. \ref{fig:Any2CapIns}), including data collection, structured caption construction, and user-centric short prompt generation.

\vspace{-4mm}
\paragraph{Step-1: Data Collection.}
\label{para:data-collection}

We begin by systematically categorizing conditions into four primary types:
1) \textbf{Spatial-wise conditions}, which focus on the structural and spatial properties of the video, e.g., \emph{depth maps}, \emph{sketches}, and \emph{video frames}.
2) \textbf{Action-wise conditions}, which emphasize motion and human dynamics in the target video, e.g., \emph{human pose}, \emph{motion}.
3) \textbf{Composition-wise conditions}, which focus on scene composition, particularly in terms of object interactions and multiple identities in the target video.
4) \textbf{Camera-wise conditions}, which control video generation from a cinematographic perspective, e.g., \emph{camera angles}, \emph{movement trajectories}. 
Since it is infeasible to encompass all possible conditions in dataset collection, we curate representative datasets under each category, specifically including depth maps, human pose, multiple identities, and camera motion.
During the data collection process, we leverage SoTA tools to construct conditions.
For instance, Depth Anything \cite{Depth-Anything-YangKH0XFZ24} is used to generate depth maps, DW-Pose \cite{DW-Pose-YangZY023} provides human pose annotations, and SAM2 \cite{SAM-2} is utilized for segmentation construction. 
In total, we collect \textbf{337K} video instances and \textbf{407K} conditions, with detailed statistics of the dataset presented in Tab. \ref{tab:collected-dataset}.

\begin{figure}[t]
\centering
\includegraphics[width=0.99\linewidth]{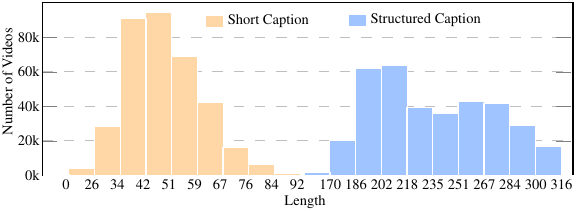}
\vspace{-3mm}
\caption{\textbf{Distribution of the short/structured caption length} (in words) in \texttt{Any2CapIns}. }
\label{fig:length-dis}
\vspace{-4mm}
\end{figure}

\vspace{-3mm}
\paragraph{Step-2: Structured Video Caption Generation.}

The granularity of a caption, specifically the detailed selection of elements they encompass, plays a critical role in guiding the model to produce videos that closely align with the users' intentions while preserving coherence and realism. 
Drawing inspiration from \cite{MiraData-Ju00YWZX0S24}, we inherit its structured caption format consisting of (1) \texttt{Dense caption}, (2) \texttt{Main object caption}, (3) \texttt{Background caption}, (4) \texttt{Camera caption}, and (5) \texttt{Style caption}.
Furthermore, the action descriptions of the subjects significantly influence the motion smoothness of the videos \cite{Koala-36M}, we explicitly incorporate the  (6) \texttt{Action caption} of subjects to form the final structured caption.
Technically, following \cite{Koala-36M}, we independently generate each caption component and subsequently integrate them to construct the final structured caption.

\vspace{-3mm}
\paragraph{Step-3: User-centric Short Prompt Generation.}

In this step, we construct short prompts from a user-centric perspective, considering how users naturally express their intentions. 
Firstly, our analysis highlights three key characteristics of user-generated prompts:
1) \textbf{Conciseness and Simplicity}, where users favor brief and straightforward wording;
2) \textbf{Condition-Dependent Omission}, whereby users often omit explicit descriptions of certain attributes (e.g., camera movement) when such conditions are already specified;
and 3) \textbf{Implicit instruction of Target Video}: where users convey their intent indirectly (e.g., specifying multiple identities without explicitly detailing their interactions).
Guided by these observations, we employ GPT-4V to infer potential user prompts under condition-specific constraints. 
Given a structured video caption, target video, and associated conditions, we apply tailored constraints to preserve condition invariance when relevant. 
We also explicitly control the length of the generated prompts to ensure conciseness. 
Finally, we conduct manual verification and filtering to further refine the dataset. 
Fig. \ref{fig:length-dis} presents the length distribution of the resulting short and structured prompts.

\begin{table}[!t]
    \centering
    \fontsize{8}{9}\selectfont
    \setlength{\tabcolsep}{1mm}
    \begin{tabular}{l cccc}
    \toprule
        {\textbf{Category}} & {\textbf{\#Inst.}} & \textbf{\#Condition}  & \textbf{\#Avg. Len.} & \textbf{\#Total Len.} \\
    \midrule
       Depth  & 182,945 & 182,945 & 9.87s & 501.44h \\
       Human Pose  & 44,644 & 44,644 & 8.38s & 108.22h  \\
       Multi-Identities & 68,255 & 138,089 & 13.01s & 246.69h  \\
       Camera Movement & 41,112 & 41,112 & 6.89s & 78.86h    \\
    \bottomrule
    \end{tabular}
    \vspace{-2mm}
    \caption{Statistics of the collected dataset across four types of conditions. \textbf{\#Inst.} means the number of instances, and \textbf{\#Condition} denotes the number of unique conditions. \textbf{\#Avg. / \#Total Len.} indicate the average and total video durations, respectively.}
    \label{tab:collected-dataset}
    \vspace{-5mm}
\end{table}

\vspace{-1mm}
\section{Any2Caption Model}
\label{sec:Any2Cap}

\begin{figure}[!t]
  \centering
   \includegraphics[width=0.99\linewidth]{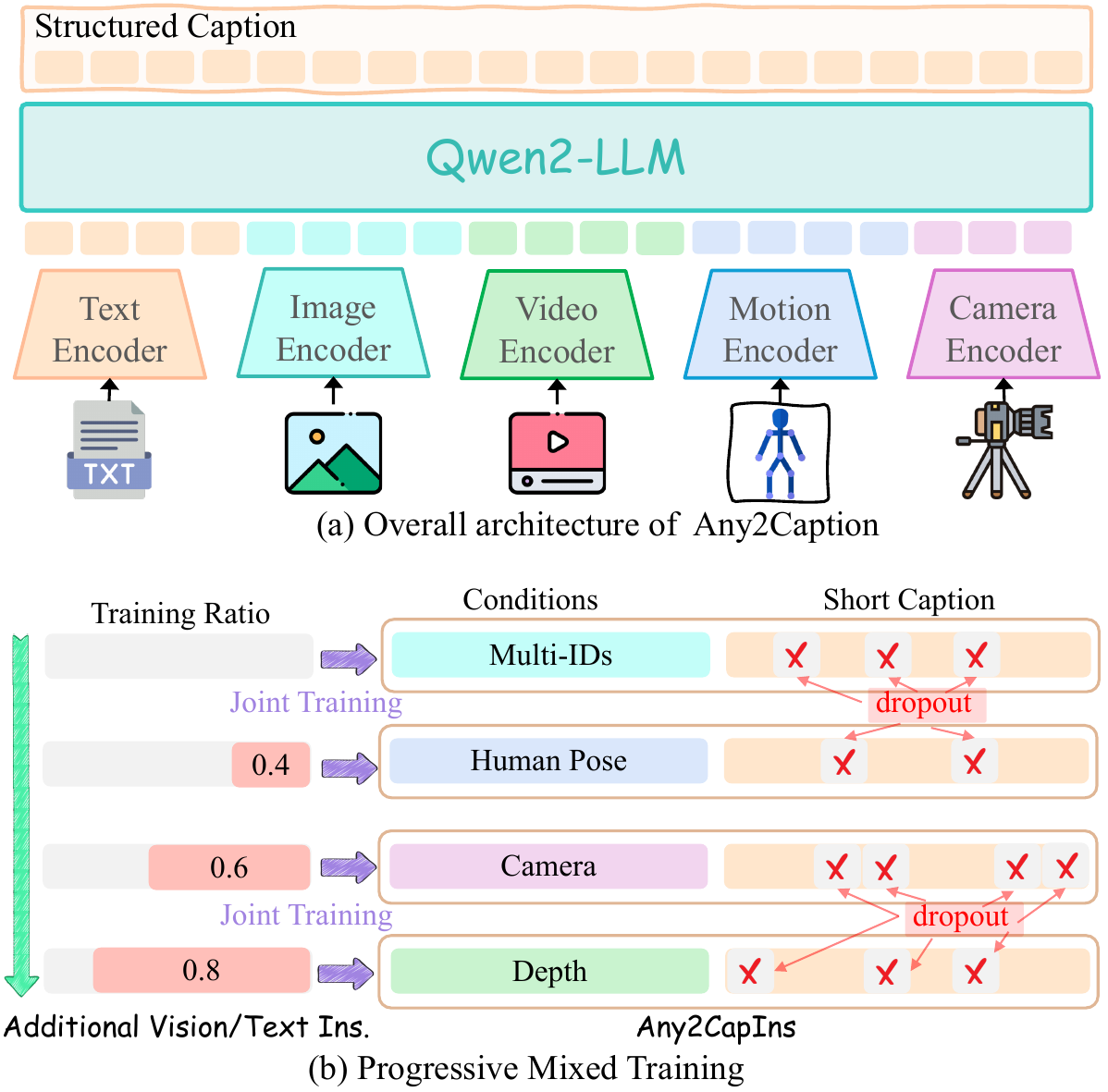}
    \vspace{-2mm}
   \caption{
   Architecture illustration of \texttt{Any2Caption} (a), where Qwen2-LLM serves as the backbone and is paired with text, image, video, motion, and camera encoders to produce structured captions.
   After alignment learning, we perform a progressive mixed training strategy (b), where additional vision/text instruction datasets are progressively added for joint training, and meanwhile, for input short caption, we adopt a random-dropout mechanism at the sentence level to enhance robustness. 
}
\label{fig:framework}
\vspace{-4mm}
\end{figure}

\vspace{-1mm}
In this section, we introduce \texttt{Any2Caption}, a novel MLLM explicitly designed to comprehensively model and interpret arbitrary multimodal conditions for controllable video caption generation, as illustrated in Fig. \ref{fig:framework}(a).
Formally, the user provides a short text prompt $T$ along with non-text conditions $C = [c_1, \cdots, c_n]$, where the non-text conditions can be either a single condition ($n=1$) or a combination of multiple conditions. 
The objective of this task is to generate a detailed and structured caption that serves as a control signal for video generation.

\vspace{-3mm}
\paragraph{Architecture.}

Similar to existing MLLMs \cite{llava-LiuLLL24,NExT-GPT-Wu0Q0C24,Video-LLaVA-LinYZCNJ024}, \texttt{Any2Caption} incorporates an image encoder $F_{\mathcal{I}}$, a video encoder $F_{\mathcal{V}}$, a motion encoder $F_{\mathcal{M}}$ and a camera encoder $F_{\mathcal{C}}\}$ to process non-text conditions.
These encoders are then integrated into an LLM backbone $F_\mathcal{LLM}$ (i.e., Qwen2-LLM) to facilitate structured video captioning. 
Specifically, we leverage a ViT-based visual encoder from Qwen2-VL as $F_{\mathcal{I}}$ and $F_{\mathcal{V}}$ for the unified modeling of images and videos, achieving effective interpretation of input conditions represented in image or video formats, such as depth maps and multiple identities.
To enable human pose understanding, we represent the extracted human pose trajectories as $\bm{H}$=$\{(x_n^{k}, y_n^{k})|k=1,\cdots,K, n=1,\cdots,N\}$, where $N$ denotes the number of video frames and $K$ is the number of keypoints.
These trajectories are then visualized within video frames to enable further processing by the motion encoder, which shares the same architectural structure and initialization as the vision encoder.
For camera motion understanding, inspired by \cite{CameraCtrl}, we introduce a camera encoder that processes a pl\"{u}cker embedding sequence $P \in \mathbb{R}^{N \times 6 \times H \times W}$, where $H$, $W$ are the height and width of the video.
This embedding accurately captures camera pose information, enabling precise modeling of camera trajectories.
Finally, in line with Qwen2-VL, we employ special tokens to distinguish non-text conditions from texts. 
Besides the existing tokens, we introduce 
\texttt{<|motion\_start|>}, \texttt{<|motion\_end|>}, \texttt{<|camera\_start|>}, \texttt{<|camera\_end|>}, 
to demarcate the start and end of human and camera pose features.

\begin{table*}[!th]
\centering
\fontsize{8.5}{10}\selectfont
\setlength{\tabcolsep}{2.7mm}
\begin{tabular}{lccccccc}
\toprule
    \multirow{2}{*}{\textbf{Category}} & \multirow{2}{*}{\textbf{Structural Integrity}} & \multicolumn{3}{c}{\textbf{Lexical Matching}} & \multicolumn{1}{c}{\textbf{Semantic Matching}} &  \multicolumn{2}{c}{\textbf{Intent Reasoning}} \\
    \cmidrule(r){3-5}\cmidrule(r){6-6}\cmidrule(r){7-8}
    & & {B-2} & {R-L} & {METER} & {BERT\textsc{Score}} & {Accuracy} & {Quality} \\
\midrule

\quad    Entire Structured Caption  & 91.25 & 54.99 & 48.63& 52.47 & 91.95 & 68.15 &  3.43 \\
\cdashline{1-8}
\quad    Dense Caption  & - & 44.24 & 42.89 & 49.51 & 92.42 & 78.47 &  3.47\\
\quad    Main Object Caption  & - & 38.54 & 47.46 & 52.48 & 92.02 & 56.28 &  2.74 \\
\quad     Background Caption  & - &44.65 & 46.73 & 48.87 & 92.90  &  69.37 &  2.69\\
\quad    Camera Caption  & - & 60.21 & 96.10 & 94.32 & 99.31 & 66.31 &  3.75 \\
\quad     Style Caption  & - & 41.71 & 47.70 & 55.9 & 93.48 & 63.75 &  3.05  \\
\quad    Action Caption  & - & 31.91 & 39.83 & 45.25 & 91.44  & 57.98 &  2.13 \\
\bottomrule
\end{tabular}
\vspace{-2mm}
\caption{Quantitative results of structured caption generation quality under four aspects: \textit{structural Integrity}, \emph{lexical matching}, \emph{semantic matching}, and \emph{intent reasoning}. 
We demonstrate the overall caption generation capability and the individual component generation performance within the structure. ``B-2'' and ``R-L'' denotes BLEU-2 and ROUGE-L, respectively.}
\label{tab:cap-gen}
\vspace{-3mm}
\end{table*}

\vspace{-3mm}
\paragraph{Training Recipes.}

To accurately interpret user generation intent under arbitrary conditions and yield structured target video captions, large-scale pretraining and instruction tuning are required. 
To this end, we adopt a two-stage training procedure:
\textbf{Stage-I: Alignment Learning.}
In this stage, as image and video encoders have been well-trained in Qwen2-VL, we only focus on aligning human pose features from the motion encoder and camera movement features with the word embeddings of the LLM backbone. 
To achieve this, we freeze the LLM and vision encoder while keeping the motion encoder trainable and optimizing it on a human pose description task.
Similarly, for camera movement alignment, we unfreeze the camera encoder and train it on a camera movement description task, ensuring that camera-related conditions are embedded into the model's latent space. 
This alignment phase establishes a strong foundation for effective representation learning for these conditions.
\textbf{Stage II: Condition-Interpreting Learning}
Building upon the aligned encoders and pretrained Qwen2-VL weights, we fine-tune the model on the Any2CapIns dataset for multimodal condition interpretation.
However, direct fine-tuning leads to catastrophic forgetting due to the fixed output structure.
To address this, we propose a progressive mixed training strategy.
Specifically, the model is first trained on a single condition to establish a strong condition-specific understanding.
As new conditions are introduced, we gradually incorporate vision-language instructions such as LLaVA-instruction \cite{llava-LiuLLL24} and Alpaca-52K \cite{alpaca}.
This stepwise strategy ensures robust multimodal condition interpretation while preventing knowledge degradation.

\section{Evaluation Suite}
\label{sec:evaluation}
\vspace{-1mm}

\vspace{-1mm}
In this section, we introduce the evaluation suite for comprehensively assessing the capability of \texttt{Any2Caption} in interpreting user intent and generating structured captions.

\vspace{-3mm}
\paragraph{Lexical Matching Score.}

To assess caption generation quality from a lexical standpoint, we employ standard evaluation metrics commonly used in image/video captioning tasks, including \texttt{BLEU} \cite{Papineni02bleu}, \texttt{ROUGE} \cite{lin-2004-rouge}, and \texttt{METEOR} \cite{banarjee2005}.
We also introduce a \texttt{Structural Integrity} score to verify whether the generated captions adhere to the required six-component format, thereby ensuring completeness.

\vspace{-3mm}
\paragraph{Semantic Matching Score.}

To evaluate the semantic alignment of generated captions, we employ BERT\textsc{Score} \cite{BERTScore}, which computes similarity by summing the cosine similarities between token embeddings, effectively capturing both lexical and compositional meaning preservation. 
Additionally, we utilize CLIP Score \cite{CLIPScore-HesselHFBC21} to assess the semantic consistency between the input visual condition and the generated videos.

\begin{table}[!t]
    \centering
    \fontsize{8}{9}\selectfont
    \setlength{\tabcolsep}{1.0mm}
    \begin{tabular}{lcccc}
    \toprule
        \multirow{2}{*}{\textbf{\makecell{Caption \\ Enrich}}} & \multicolumn{1}{c}{\textbf{Text}} & \multicolumn{3}{c}{\textbf{Video Generation}} \\
        \cmidrule(r){2-2}\cmidrule(r){3-5}
          & CLIP-T$\uparrow$ & Smoothness$\uparrow$ & Aesthetic$\uparrow$ & Integrity$\uparrow$ \\
    \midrule
    Short Cap. & 18.31 & 93.46 & 5.32 & 55.39 \\
    Short Cap.  &  \multirow{2}{*}{19.19} &  \multirow{2}{*}{93.41} &  \multirow{2}{*}{5.41} &  \multirow{2}{*}{54.91}\\
    \quad w/ Condition Cap. & & & & \\ 
    \rowcolor{lightgrey} Structured Cap. & \textbf{19.87} & \textbf{94.38} & \textbf{5.46} & \textbf{57.47} \\
    \bottomrule
    \end{tabular}
    \vspace{-2mm}
    \caption{Quantitative results comparing short caption, short caption combined with condition caption, and structured caption for multi-identity video generation.
}
\label{tab:cap_comparison}
\vspace{-4mm}
\end{table}

\vspace{-3mm}
\paragraph{Intent Reasoning Score.}

Evaluating structured captions typically focuses on their linguistic quality, yet it rarely assesses whether the model truly understands user intent and accurately reflects it across aspects such as style, emotion, and cinematic language.
To bridge this gap and draw inspiration from \cite{AuroraCap}, we introduce the Intent Reasoning Score (IR\textsc{Score}), a novel quantitative metric that leverages LLMs to assess whether generated captions accurately capture user intentions. 
IR\textsc{Score} identifies user-specified control factors, then decomposes the generated caption into QA pairs aligned with these factors. 
The framework has four steps: (1) \textbf{User Intention Extraction:} Analyze provided conditions to categorize user intent into six aspects: subject, background, movement, camera, interaction, and style.
(2) \textbf{Ground-Truth QA Pair Construction:} Formulate aspect-specific QA pairs with defined requirements (e.g., for subject-related attributes, emphasize object count, appearance). 
(3) \textbf{Predicted Answer Generation:} Prompt GPT-4V to parse the predicted caption and generate answers based solely on it. 
(4) \textbf{Answer Evaluation:} Following \cite{VideoChat-abs-2305-06355}, GPT-4V outputs two scores (correctness and quality) for each answer, which are then averaged across all QA pairs. 
More details are in Appendix $\S$\ref{app:eval_suite}.

\vspace{-3mm}
\paragraph{Video Generation Quality Score.}

The primary objective of generating structured captions is to enhance the video generation quality and controllability. 
Therefore, we adopt a series of metrics to assess the quality of videos generated based on structured captions.
Following \cite{VBench-HuangHYZS0Z0JCW24,MiraData-Ju00YWZX0S24}, we evaluate video generation quality across four key dimensions:
motion smoothness, dynamic degree, aesthetic quality, and image integrity.
To further verify adherence to specific non-text conditions, we use specialized metrics: RotErr, TransErr, and CamMC \cite{CameraCtrl} for camera motion accuracy; Mean Absolute Error (MAE) for depth consistency \cite{SparseCtrl-GuoYRALD24}; DINO-I \cite{DreamBooth-RuizLJPRA23}, CLIP-I \cite{DreamBooth-RuizLJPRA23} Score to evaluate identity preservation under multiple identities, and Pose Accuracy (Pose Acc.) \cite{MaHCWC0C24} to access the alignment in the generated videos.

\begin{table}[!t]
\centering
\fontsize{8}{9}\selectfont
\setlength{\tabcolsep}{0.5mm}
\begin{tabular}{l cc cccc}
\toprule
    \multirow{2}{*}{\textbf{\makecell{Training \\ Strategy}}} & \multicolumn{2}{c}{\textbf{Caption}} & \multicolumn{3}{c}{\textbf{Vieo Generation}} \\
    \cmidrule(r){2-3}\cmidrule(r){4-6}
    & {B-2}$\uparrow$ & {Accuracy}$\uparrow$ & {Smoothness}$\uparrow$ & {Dynamics}$\uparrow$ & {Aesthetic}$\uparrow$ \\
\midrule
\rowcolor{lightgrey} Any2Caption & 47.69 & 67.35 & \textbf{94.60} & \textbf{17.67} & \textbf{5.53} \\
\cdashline{1-6}
\quad w/o Two-Stage & 33.70 & 51.79  & 93.31 & 16.36 & 5.50 \\
\quad w/o Dropout & \textbf{49.24} & \textbf{69.51} & 94.16 & 14.54 & 5.51  \\

\bottomrule
\end{tabular}
\vspace{-2mm}
\caption{
Ablation study on training strategy. 
``w/o stage'' means alignment learning is not applied during training, while ``w/o Dropout'' denotes that short captions are not randomly dropped.
}
\label{tab:ablation}
\vspace{-4mm}
\end{table}

\begin{figure*}[!t]
\centering
\includegraphics[width=0.99\textwidth]{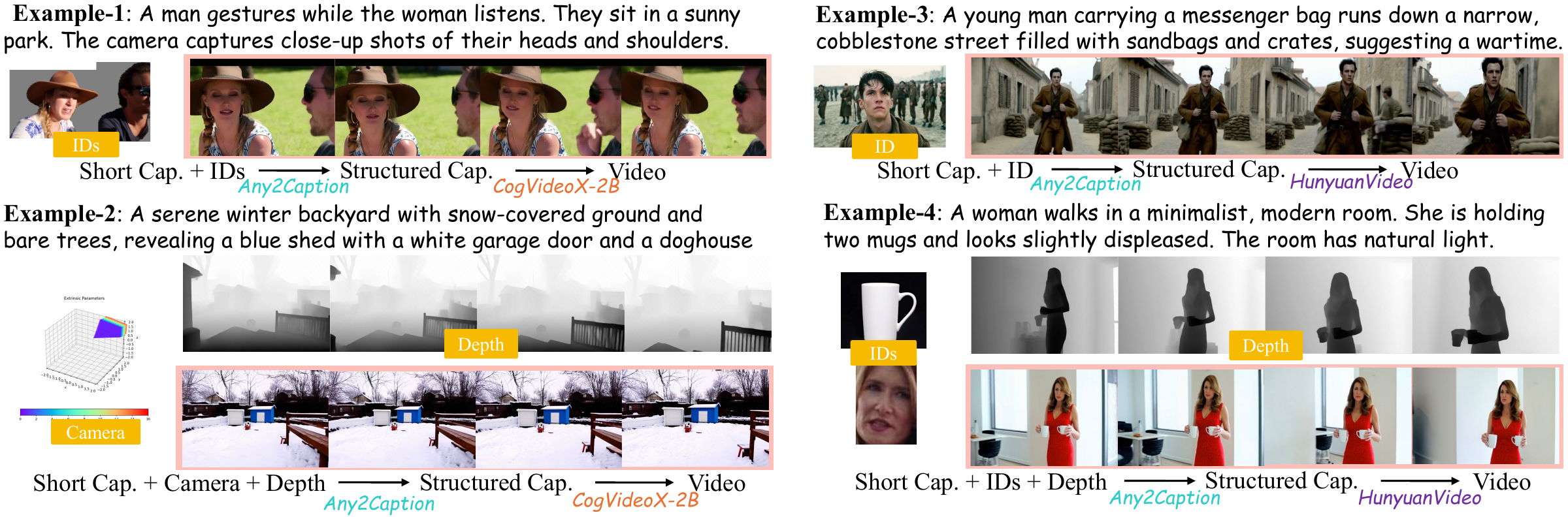}
\vspace{-2mm}
\caption{
    Illustrations of \colorbox{pinkBG}{generated videos} where only the structured captions yielded by \textit{\textcolor{titlegreen}{Any2Caption}} are fed into the \textit{\textcolor{cogvideo}{CogVideoX-2B}} (Left), and \textit{\textcolor{hunyuan}{HunyuanVideo}} (Right). 
    We can observe that some key features of the input identity images, such as the background and main object, can be accurately visualized in the generated videos. 
}
\vspace{-2mm}
\label{fig:t2v-cap}
\end{figure*}

\section{Experiments}

\subsection{Setups}

\vspace{-1mm}

\paragraph{Dataset.}

We manually construct 200 test cases for each type of condition (i.e., \textit{depth}, \textit{human pose}, \textit{multiple identities}, \textit{camera}, and \textit{compositional conditions}) to evaluate the model's performance. 
Additionally, we assess the model on publicly available benchmarks (e.g., \cite{StyleMaster, Ctrl-Adapter,ZhouTFFS18}). For further details, please refer to the Appendix $\S$\ref{app:setups}.

\vspace{-2mm}
\paragraph{Implementation Details.}

We leverage Qwen2VL-7B \cite{Qwen2-VL} as the backbone of our model, which supports both image and video understanding. 
The human pose in the input conditions is encoded and processed in the video format. 
The camera encoder adopts the vision encoder architecture with the following settings: the input channel of 96, the patch size of 16, the depth of 8, and 8 attention heads. 
During training, to simulate the brevity and randomness of user inputs, we randomly drop sentences from the short caption with a dropout rate of 0.6; a similar dropout strategy is applied to non-textual conditions. 
We conducted the training on 8$\times$80G GPUs. 
For further details on the training parameters for each stage, please refer to the Appendix $\S$\ref{app:setups}.

\begin{table*}[!th]
\centering
\fontsize{8}{9}\selectfont
\setlength{\tabcolsep}{0.5mm}
\begin{tabular}{l c ccc cc c c cccc}
\toprule
\multirow{2}{*}{\textbf{Model}} & \multicolumn{1}{c}{\textbf{Text}} & \multicolumn{3}{c}{\textbf{Camera}} & \multicolumn{2}{c}{\textbf{Identities}} & \multicolumn{1}{c}{\textbf{Depth}} & \multicolumn{1}{c}{\textbf{Human Pose}} & \multicolumn{4}{c}{\textbf{Overall Quality}} \\
\cmidrule(r){3-5}\cmidrule(r){6-7}\cmidrule(r){8-8}\cmidrule(r){9-9}\cmidrule(r){10-13}     
& CLIP-T$\uparrow$ & RotErr$\downarrow$ & TransErr$\downarrow$ & CamMC$\downarrow$ & DINO-I$\uparrow$ & CLIP-I$\uparrow$ & MAE$\downarrow$ & Pose Acc.$\uparrow$ & Smoothness$\uparrow$ & Dynamic$\uparrow$ & Aesthetic$\uparrow$ &  Integrity$\uparrow$ \\ 
\midrule
\multicolumn{12}{c}{$\bullet$ \textbf{Camera to Video}} \\
MotionCtrl \cite{MotionCtrl-WangYWLCXLS24} & 19.67 & 1.54 & 4.49 & 4.80 & - & - & - & - & 96.13 &  9.75 & 5.40 & 73.69 \\
\rowcolor{lightgrey} \quad + Structured Cap. & \textbf{20.16} & \textbf{1.45} & \textbf{4.37} &  \textbf{4.78} & - & - & - & - & \textbf{96.16} &  \textbf{11.43} & \textbf{5.71} & \textbf{74.63}  \\
\cdashline{1-13}
CameraCtrl \cite{CameraCtrl}  & 18.89 & 1.37 & 3.51 & 4.78 & - & - & -  & - & 94.11 & 12.59 & 4.26  & 71.84  \\
\rowcolor{lightgrey} \quad + Structured Cap. & \textbf{21.70} & \textbf{0.94} & \textbf{2.97} & \textbf{4.37} & - & - & - & - & \textbf{95.16} & \textbf{13.72} & \textbf{4.66}  & \textbf{72.47} \\

\hline

\multicolumn{12}{c}{$\bullet$ \textbf{Depth to Video}} \\
Ctrl-Adapter \cite{Ctrl-Adapter} &  20.37 & - & -  &  - & - & - & 25.63 & - & 94.53 & \textbf{20.73} & 4.63 &  46.98 \\
\rowcolor{lightgrey} \quad + Structured Cap. &  \textbf{23.30} & - & -  &  - & - & - & \textbf{21.87} & - & \textbf{95.54} & {15.14} & \textbf{5.31}  &  \textbf{54.20} \\
\cdashline{1-13}

ControlVideo \cite{ControlVideo}  & 22.17 & - & -  &  - & - & - & 30.11 & - &  92.88 & 5.94 & 5.29  & 63.85  \\
\rowcolor{lightgrey} \quad + Structured Cap. &  \textbf{24.18} & - & -  &  - & - & - & \textbf{23.92} & - & \textbf{94.47} & \textbf{18.27} & \textbf{9.77}  & \textbf{66.28}  \\
\hline

\multicolumn{12}{c}{$\bullet$ \textbf{Identities to Video}} \\
ConceptMaster \cite{ConceptMaster}  & 16.04 &  - & -  &  - &  36.37 & 65.31 & - &  - & 94.71 & 8.18 & 5.21  & 43.68  \\
\rowcolor{lightgrey} \quad + Structured Cap.  & \textbf{17.15} &  - & -  &  - &  \textbf{39.42} & \textbf{66.74} & - &  - & \textbf{95.05} & \textbf{10.14} & \textbf{5.68}  & \textbf{49.73}    \\
\hline

\multicolumn{12}{c}{$\bullet$ \textbf{Human Pose to Video}} \\
FollowYourPose \cite{MaHCWC0C24} & 21.11 &  - & -  &  - & - & -  &  - & 30.47 & 91.71  &  14.29 & 4.95 & \textbf{58.84}  \\
\rowcolor{lightgrey} \quad + Structured Cap.  & \textbf{21.39} &  - & -  &  - & - & - & - &  \textbf{31.59} & \textbf{92.87} & \textbf{16.47} & \textbf{5.88} & {56.30}  \\
\bottomrule
\end{tabular}
\vspace{-2mm}
\caption{
Performance comparison on four types of conditions (e.g., \textit{camera}, \textit{depth}, \textit{identities}, and \textit{human pose}) between directly using short captions and integrating structured captions under various video quality evaluation metrics. Better results are marked in \textbf{bold}.
}
\label{tab:video-gen}
\vspace{-4mm}
\end{table*}

\subsection{Experimental Results and Analyses}
\vspace{-1mm}

In this section, we present the experimental results and provide in-depth analyses to reveal how the system advances.
Following this, we try to ground the answers for the following six key research questions.

\vspace{-3mm}
\paragraph{RQ-1: How well is the structured caption generation quality?}

We first evaluate whether our proposed model could accurately interpret user intent and generate high-quality structured captions. 
From a caption-generation perspective, we compare the predicted captions with gold-standard captions across various metrics (refer to Tab. \ref{tab:cap-gen}). 
We observe that our model successfully produces the desired structured content, achieving 91.25\% in structural integrity. 
Moreover, it effectively captures the key elements of the gold captions, attaining a ROUGE-L score of 48.63 and a BERT\textsc{Score} of 91.95. 
Notably, the model demonstrates the strongest performance in interpreting camera-related details compared to other aspects.
Finally, regarding user intent analysis, we found that the model reliably incorporated user preferences into its structured outputs.

To further showcase the model's capacity to understand and leverage input conditions, we directly feed the structured captions—derived from our model's interpretation—into downstream text-to-video generation systems (e.g., CogvideoX-2B \cite{CogVideoX} and Hunyuan \cite{kong2024hunyuanvideo}), as illustrated in Fig. \ref{fig:t2v-cap}. 
Even without explicit visual conditions (e.g., identities, depth, or camera movement), the resulting videos clearly align with the input prompts, such as the hat's color and style, or the woman's clothing texture in Example 1, indicating that our structured captions successfully capture intricate visual details. 
In particular, the model is able to accurately grasp dense conditions, such as depth sequences or compositional requirements in Example 4, ultimately enabling controllable video generation. 
While certain fine-grained elements may not be exhaustively described in text, resulting in occasional discrepancies with the actual visual content, the overall controllability remains robust.

\vspace{-3mm}
\paragraph{RQ-2: Is the video generation quality enhanced with structured caption?}

Here, we investigate whether integrating structured captions consistently improves controllable video generation in multiple methods.
We explore the impact of adding camera, depth, identities, and human pose conditions to various controllable video generation methods. 
As shown in Tab.~\ref{tab:video-gen}, all tested models exhibit consistent gains in overall video quality, such as smoothness and frame fidelity, after incorporating structured captions, without requiring any changes to the model architectures or additional training. 
Moreover, these models show enhanced adherence to the specified conditions, suggesting that our generated captions precisely capture user requirements and lead to more accurate, visually coherent video outputs.
More examples can be found in the Appendix $\S$\ref {app:vis}.

\begin{table*}[!th]
\centering
\fontsize{8}{9}\selectfont
\setlength{\tabcolsep}{0.8mm}
\begin{tabular}{l c ccc cc c cccc}
\toprule
\multirow{2}{*}{\textbf{Compositional  Condition}} & \multicolumn{1}{c}{\textbf{Text}} & \multicolumn{3}{c}{\textbf{Camera}} & \multicolumn{2}{c}{\textbf{Identities}} & \multicolumn{1}{c}{\textbf{Depth}}  & \multicolumn{4}{c}{\textbf{Overall Quality}} \\
\cmidrule(r){2-2}\cmidrule(r){3-5}\cmidrule(r){6-7}\cmidrule(r){8-8}\cmidrule(r){9-12}     
& CLIP-T$\uparrow$ & RotErr$\downarrow$ & TransErr$\downarrow$ & CamMC$\downarrow$ & DINO-I$\uparrow$ & CLIP-I$\uparrow$ & MAE$\downarrow$ & Smoothness$\uparrow$ & Dynamic$\uparrow$ & Aesthetic$\uparrow$ &  Integrity$\uparrow$ \\ 
\midrule
 Camera+Identities & 14.81 & 1.37 & \textbf{4.04} & {4.24} & 25.63 & 64.14 & - & \textbf{94.43} & 28.87  & 4.99 & 59.81 \\
\rowcolor{lightgrey} \quad + Structured Cap. & \textbf{19.03} & \textbf{1.30} & 4.36 & \textbf{4.03} & \textbf{26.75} & \textbf{68.45} & -  & 94.38 & \textbf{34.99} & \textbf{5.25} & \textbf{63.02}\\
\cdashline{1-12}

Camera+Depth  & 20.80 & 1.57 & \textbf{3.88} & \textbf{4.77} & - & - & 32.15  & 95.36 & \textbf{30.12} & 4.82  &  63.90 \\
\rowcolor{lightgrey} \quad + Structured Cap. & \textbf{21.19} & \textbf{1.49} & 4.41 & 4.84 & - & - & \textbf{25.37} & \textbf{95.40} & 30.10 & \textbf{4.96} & \textbf{65.05} \\
\cdashline{1-12}

Depth+Identities &  20.01 & - & - & - & 35.24 & 57.82 & \textbf{23.00}  & \textbf{93.15} & 32.21 & 4.96  &  \textbf{61.21} \\
\rowcolor{lightgrey} \quad + Structured Cap. & \textbf{20.76} & - & - & - & \textbf{36.25} & \textbf{63.48} & {24.78}  & 92.50 & \textbf{36.43} & \textbf{5.18}  & {60.81 }\\
\cdashline{1-12}

Camera+Identities+Depth & 18.49 & 2.05 & 7.74 & 8.47  & 35.86 & 64.25 & 18.37   & 92.02 & 30.09 & 3.91  & 60.62 \\
\rowcolor{lightgrey} \quad + Structured Cap. & \textbf{19.52} &  \textbf{1.57} & \textbf{7.74} & \textbf{8.20} & \textbf{38.74} & \textbf{64.37} & \textbf{17.41}  & \textbf{93.03} & \textbf{32.81} & \textbf{4.99}  & \textbf{ 61.22}  \\

\bottomrule
\end{tabular}
\vspace{-2mm}
\caption{
Quantitative comparison of structured captions when handling compositional conditions. Better results are marked in \textbf{bold}.
}
\label{tab:compositional-con}
\vspace{-3mm}
\end{table*}

\vspace{-3mm}
\paragraph{RQ-3: Is the structured caption necessary?}
We examine whether a structured caption design is essential. 
We compare our structured caption approach with a simpler method, where we first caption the input condition (e.g., multiple identity images) and then concatenate that caption with the original short prompt, as shown in Tab. \ref{tab:cap_comparison}. 
Our results indicate that merely appending the condition's caption to the short prompt can reduce video smoothness and image quality. 
One likely reason is that the identity images may contain extraneous details beyond the target subject, potentially conflicting with the original prompt and causing inconsistencies.
Consequently, controllability in the final output is compromised. 
In contrast, our structured caption method accurately identifies the target subject and augments the prompt with relevant information, yielding more controllable video generation (cf. $\S$\ref{app:vis}).

\begin{figure}[!t]
\centering
\includegraphics[width=0.95\linewidth]{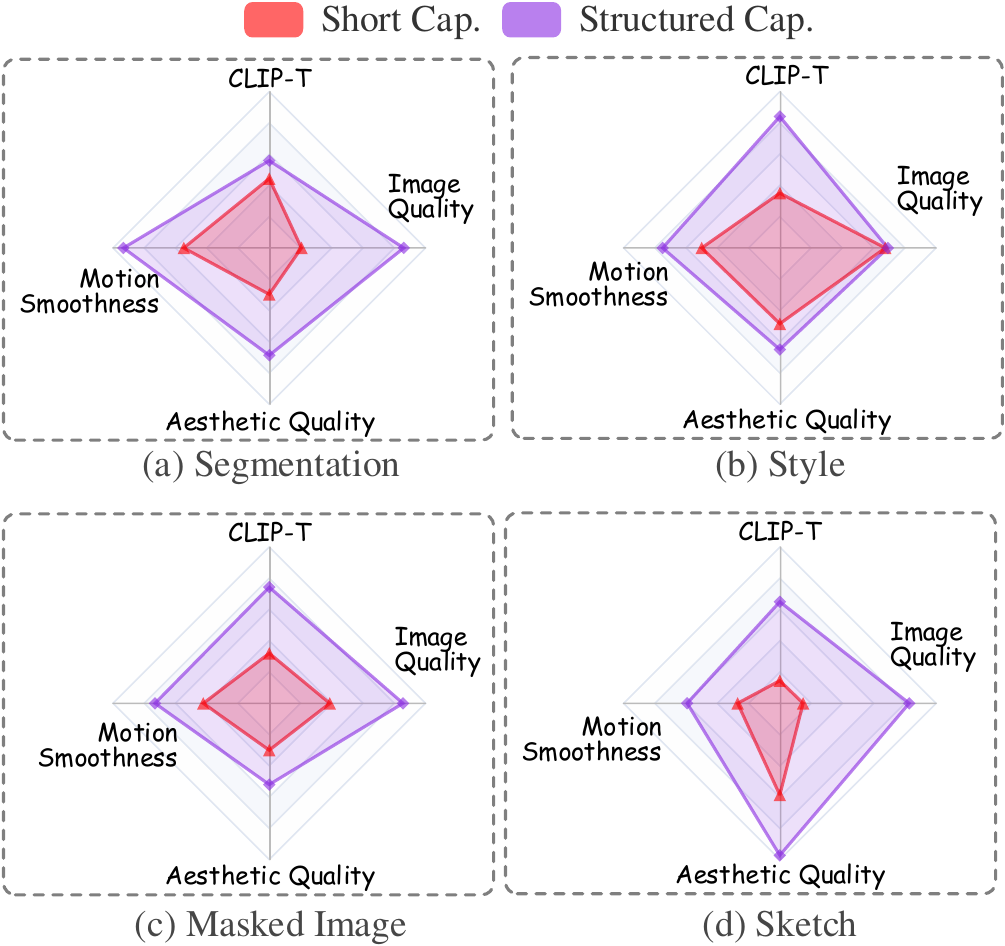}
\vspace{-2mm}
\caption{
Quantitative results on unseen conditions (i.e., \textit{segmentation} \cite{VideoComposer-WangYZCWZSZZ23}, \textit{style} \cite{StyleMaster}, \textit{masked image} \cite{VideoComposer-WangYZCWZSZZ23}, and \textit{sketch} \cite{VideoComposer-WangYZCWZSZZ23}) when using short and structured captions, respectively.
}
\vspace{-4mm}
\label{fig:unseen}
\end{figure}

\vspace{-4mm}
\paragraph{RQ-4: How effective is the training strategy?}

Next, we investigate the contribution of the training mechanism, and the results are shown in Tab. \ref{tab:ablation}.
During training, we employ a two-stage training approach, consisting of alignment learning followed by instruction-tuning. 
When alignment learning is omitted, and the model proceeds directly to instruction tuning, both captioning and video generation performance degrade significantly. 
A possible explanation is that bypassing alignment learning disrupts the encoder's adaptation process, which has been aligned to the LLM backbone, leading to suboptimal results in subsequent stages.
Additionally, we compare the performance of the model without the dropout mechanism. 
Although removing dropout yields a marked improvement in captioning quality, the benefit to video generation is marginal. 
This suggests that without dropout, the model may rely on shortcuts from the input captions rather than fully understanding the underlying intent, thereby increasing the risk of overfitting.

\vspace{-3mm}
\paragraph{RQ-5: How well does the model perform on compositional conditions?}

We examine the impact of structured captions under compositional conditions. 
As shown in Tab. \ref{tab:compositional-con}, we compare the combined camera, identities, and depth on our customized model and observe that structured captions consistently enhance its performance. 
Moreover, from the Example 2 and 4 in Fig. \ref{fig:t2v-cap}, our model demonstrates a thorough understanding of the interactions among various conditions, for instance, capturing a woman's hair color and the position of a mug, accurately guiding the production of videos that align with the specified requirements.
This finding further highlights that our approach can automatically equip existing T2V models with the ability to handle compositional conditions without requiring additional training.

\vspace{-3mm}
\paragraph{RQ-6: How well is the generalization capability of Any2Caption?}

Finally, we investigate the model's generalization ability by evaluating its performance on ``unseen'' conditions, including \textit{style}, \textit{segmentation}, \textit{sketch}, and \textit{masked images}. 
As demonstrated in Fig. \ref{fig:unseen}, the structured captions generated by our model consistently enhance existing T2V frameworks, offering benefits such as increased motion smoothness, aesthetic quality, and more accurate generation control. 
We attribute these advantages to two primary factors: the strong reasoning capabilities of our MLLM backbone and our training strategy, i.e., progressive mixed training,  which leverages existing vision and text instructions for fine-tuning while mitigating knowledge forgetting, thereby ensuring robust generalization.

\section{Conclusion}

In this work, we focus on addressing the challenge of more accurately interpreting user generation intention from any condition for controllable video generation.
We introduce \texttt{Any2Caption}, a framework that decouples multimodal condition interpretation from video synthesis.  
Built based on an MLLM, \texttt{Any2Caption} converts diverse inputs into dense captions that drive high-quality video generation.  
We further present \texttt{Any2CapIns}, a large-scale dataset for effective instruction tuning.  
Experiments show that our method improves controllability and video quality across various backbones.

{
    \small
    \bibliographystyle{ieeenat_fullname}
    \bibliography{main}
}

\clearpage
\appendix

\section*{Appendix Overview}

The appendix presents more details and additional results not included in the main paper due to page limitations. The list of items included is:

\begin{compactitem}
\item Limitation in Section $\S$\ref{app:limitation}.
\item Ethic Statement results in Section $\S$\ref{app:ethic_statement}.
\item Extended Related Work in Section $\S$\ref{sec:extend_related_work}.
\item Extended Dataset Construction Details in Section $\S$\ref{app:dataset_details}.
\item More Statistics Information of IR\textsc{Score} in Section $\S$\ref{app:eval_suite}.
\item  Detailed Setups in Section $\S$\ref{app:setups}.
\item  Extended Experiment Results and Analyses in Section $\S$\ref{app:experiments}.
\end{compactitem}

\section{Limitation}
\label{app:limitation}

Despite the advancement of our proposed framework, several limitations may remain:

Firstly, the diversity of annotated data is constrained by the capabilities of the current annotation tools, which may limit the variety of generated content. Moreover, the scarcity of real-world data introduces potential domain gaps, reducing the model’s generalizability in practical scenarios.

Secondly, due to inherent model limitations, hallucinations may occur, resulting in inaccurate structured captions and consequently degrading the quality of generated videos. A possible direction to mitigate this issue is to develop an end-to-end approach that jointly interprets complex conditions and handles video generation.

Lastly, the additional condition-understanding modules inevitably increase inference time. However, our empirical results suggest that the performance gains from these modules are substantial, and future work may explore more efficient architectures or optimization techniques to balance speed and accuracy.

\section{Ethic Statement}
\label{app:ethic_statement}

This work relies on publicly available datasets and manually constructed datasets, ensuring that all data collection and usage adhere to established privacy standards. We recognize that automatic annotation processes may introduce biases, and we have taken measures to evaluate and mitigate these biases. Nonetheless, we remain committed to ongoing improvements in this area.

By enhancing video generation capabilities, \texttt{Any2Caption} could inadvertently facilitate negative societal impacts, such as the production of deepfakes and misinformation, breaches of privacy, or the creation of harmful content. We, therefore, emphasize the importance of strict ethical guidelines, robust privacy safeguards, and careful dataset curation to minimize these risks and promote responsible research practices.

\section{Extended Related Work}
\label{sec:extend_related_work}

\subsection{Text-to-Video Generation}

The development of video generation models has progressed from early GAN- and VAE-based approaches \cite{BrooksHAWAL0EK22,WangBBD20,HaimFGSBDI22,ChuXMLT20,GurBW20} to the increasingly popular diffusion-based methods \cite{BlattmannRLD0FK23,abs-2311-15127,ZhangXZFC24,QingZWWWZGS24}. 
Among these, diffusion-in-transformer (DiT) architectures, such as OpenAI's Sora \cite{sora} and HunyuanVideo \cite{kong2024hunyuanvideo}, have demonstrated remarkable performance, producing photorealistic videos over extended durations. 
Controllable video generation \cite{abs-2411-04928,abs-2408-17424,abs-2411-18281,CameraCtrl} has become an essential aspect of this field. 
Initially, research efforts centered predominantly on text-to-video generation \cite{Make-A-Video,CogVideo-Hong0ZL023}, which remains the most common approach. 
However, relying solely on text prompts can be insufficient for accurately capturing user intent, prompting exploration into other conditioning inputs such as static images \cite{Tune-A-Video-WuGWLGSHSQS23,I2V-Adapter-GuoZHGDWZLHZHM24}, user sketches \cite{ControlVideo,VideoComposer-WangYZCWZSZZ23}, human poses \cite{PoseCrafter-ZhongZYYZL24,MaHCWC0C24,DreamPose-KarrasHWK23}, camera perspectives \cite{CamI2V,CameraCtrl}, and even additional videos \cite{KaraKYRY24,DengWZTT24,I2VGen-XL}. 
Given this diversity of potential conditions, unifying them into a single ``any-condition'' video generation framework is highly valuable.

\subsection{Controllable Video Generation}

Recent methods like VideoComposer \cite{VideoComposer-WangYZCWZSZZ23}, Ctrl-Adapter \cite{Ctrl-Adapter}, and ControlVideo \cite{ControlVideo} have investigated any-condition video generation. 
Nevertheless, they still struggle with comprehensive controllability due to the complexity of multiple modalities and the limited capacity of standard diffusion or DiT encoders to interpret them. 
Inspired by the strong multimodal reasoning capabilities of modern MLLMs \cite{llava-LiuLLL24,Video-LLaVA-LinYZCNJ024,Qwen2-VL}, we propose leveraging an MLLM to consolidate all possible conditions into a unified reasoning process, producing structured dense captions as inputs to a backbone Video DiT. 
SoTA DiT models already exhibit the capacity to interpret dense textual descriptions, as long as the input captions are sufficiently detailed in depicting both the scene and the intended generation goals. 
Building on this, our MLLM-based condition encoder directly addresses the comprehension bottleneck, theoretically enabling higher-quality video generation. 
To our knowledge, this work is the first to develop an MLLM specifically tailored for any-condition video generation. Because the caption-generation mechanism is decoupled from DiT, our proposed \texttt{Any2Caption} can be integrated into existing DiT-based methods without additional retraining.

\subsection{Video Captioning}

Our approach is closely related to video recaptioning research, as our MLLM must produce dense captions based on the given conditions. 
In text-to-video settings, prior work \cite{InstanceCap,CogVideoX,abs-2407-02371,IslamHYNTB24} has demonstrated the benefits of recaptioning videos to obtain more detailed textual annotations, thereby improving the training of longer and higher-quality video generation via DiT. 
ShareGPT4Video \cite{ShareGPT4Video}, for example, employs GPT-4V \cite{Gpt-4v} to reinterpret video content and produce richer captions. 
MiraData \cite{MiraData-Ju00YWZX0S24} introduces structured dense recaptioning, while InstanceCap \cite{InstanceCap} focuses on instance-consistent dense recaptioning. 
Although we also pursue structured dense captions to enhance generation quality, our method diverges fundamentally from these previous approaches. 
First, because DiT models are already sufficiently powerful, we directly adopt an off-the-shelf Video DiT without incurring the substantial cost of training it with dense captions. 
Instead, we train an MLLM as an any-condition encoder at a comparatively lower cost; in the text-to-video scenario, for instance, we only need to train on pairs of short and dense captions, which are far easier and more abundant to obtain. 
Second, prior methods that recapturing the entire video risk introducing noise or even hallucinated content due to the current limitations of MLLMs in video understanding, potentially undermining DiT training quality, whereas our framework avoids this issue. 
Most importantly, while these approaches may rely on dense-caption-trained DiT models, real-world inference often involves very concise user prompts, creating a mismatch that can diminish final generation quality.

\subsection{Multimodal Large Language Models}

Recent advances in Large Language Models (LLMs) \cite{Qwen2} have catalyzed a surge of interest in extending their capabilities to multimodal domains \cite{Video-LLaVA-LinYZCNJ024,LLaMA-VID-LiWJ24,Mini-Gemini-abs-2403-18814,WuZXL00WZLL0QD24,OMG-LLaVA-ZhangL0YWJLY24}. 
A number of works integrate a vision encoder (e.g., CLIP \cite{llava-LiuLLL24}, DINOv2\cite{Cambrian-1-TongBWWIAYYMWPF24}, OpenCLIP\cite{Mini-Gemini-abs-2403-18814}) with an LLM, often through a lightweight ``connector'' module (e.g., MLP \cite{llava-LiuLLL24}, Q-former \cite{BLIP-2-0008LSH23}), enabling the model to process image and video inputs with minimal additional training data and parameters \cite{llava-LiuLLL24,NExT-GPT-Wu0Q0C24,Video-LLaVA-LinYZCNJ024}. 
These approaches have demonstrated promising performance on tasks such as image captioning, visual question answering, and video understanding. 
Beyond purely visual data, some researchers have investigated broader modalities, such as 3D motion \cite{MotionLLM-abs-2405-20340,MotionGPT-JiangCLYYC23} or audio \cite{Qwen2-Audio, NExT-GPT-Wu0Q0C24}, thereby expanding the application range of multimodal LLMs.
Despite these advances, most existing MLLMs focus on a limited set of visual modalities and do not readily accommodate more specialized inputs like human pose or camera motion. 
This gap restricts their ability to handle more diverse and complex conditions in the controllable video generation field. 
In contrast, our work targets a broader spectrum of modalities, aiming to build a single model capable of interpreting and unifying image, video, human pose, and camera conditions. 
Specifically, we augment an existing MLLM with dedicated encoders for motion and camera features, thereby equipping the model to process arbitrary multimodal conditions and facilitate controllable video generation.

\begin{figure*}[!t]
\centering
\includegraphics[width=0.99\textwidth]{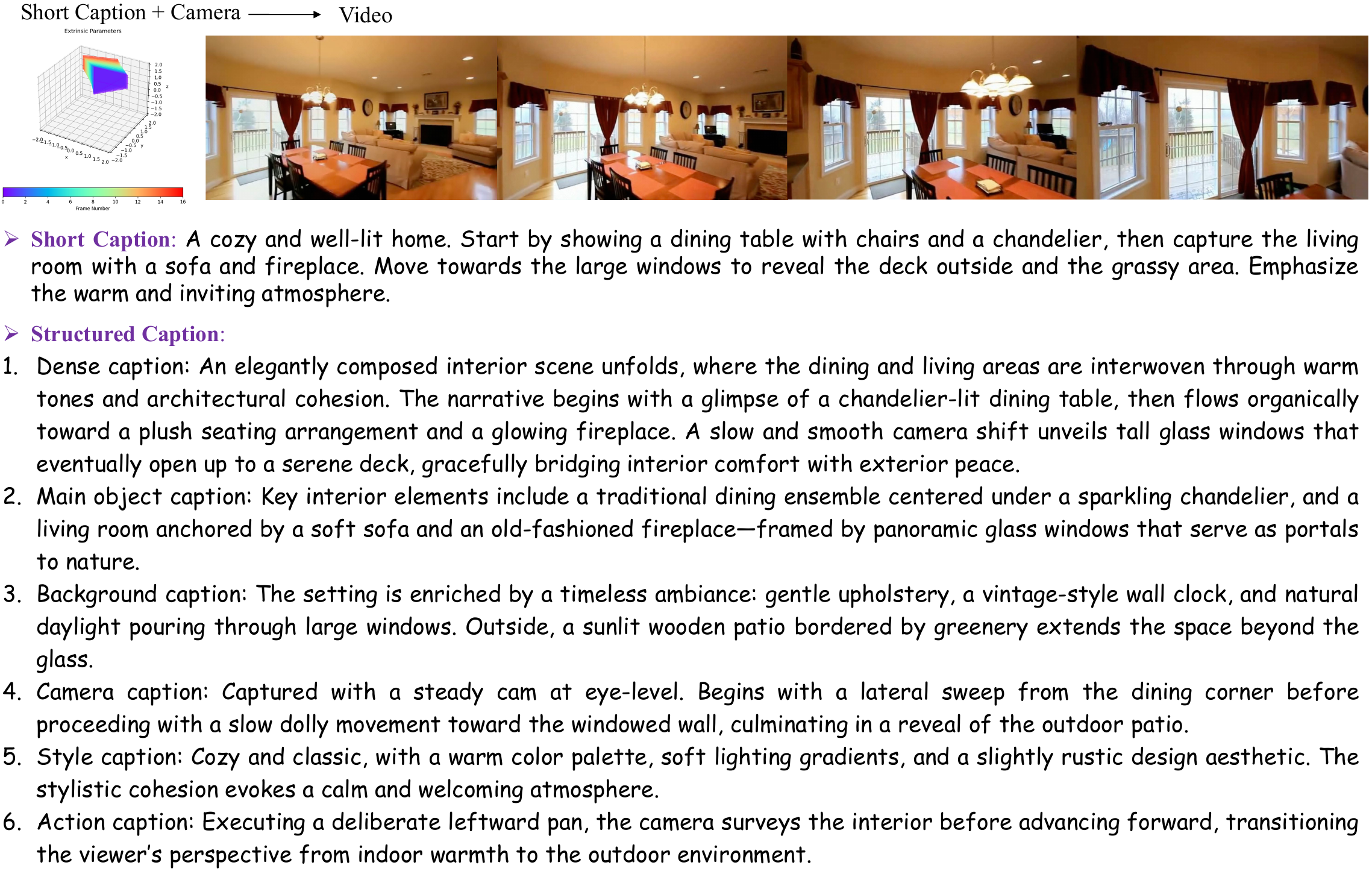}
\vspace{-2mm}
\caption{
    Illustrations of constructed short and structured captions under the camera-to-video generation. 
}
\vspace{-3mm}
\label{fig:caption-case-1}
\end{figure*}

\begin{figure*}[!t]
\centering
\includegraphics[width=0.99\textwidth]{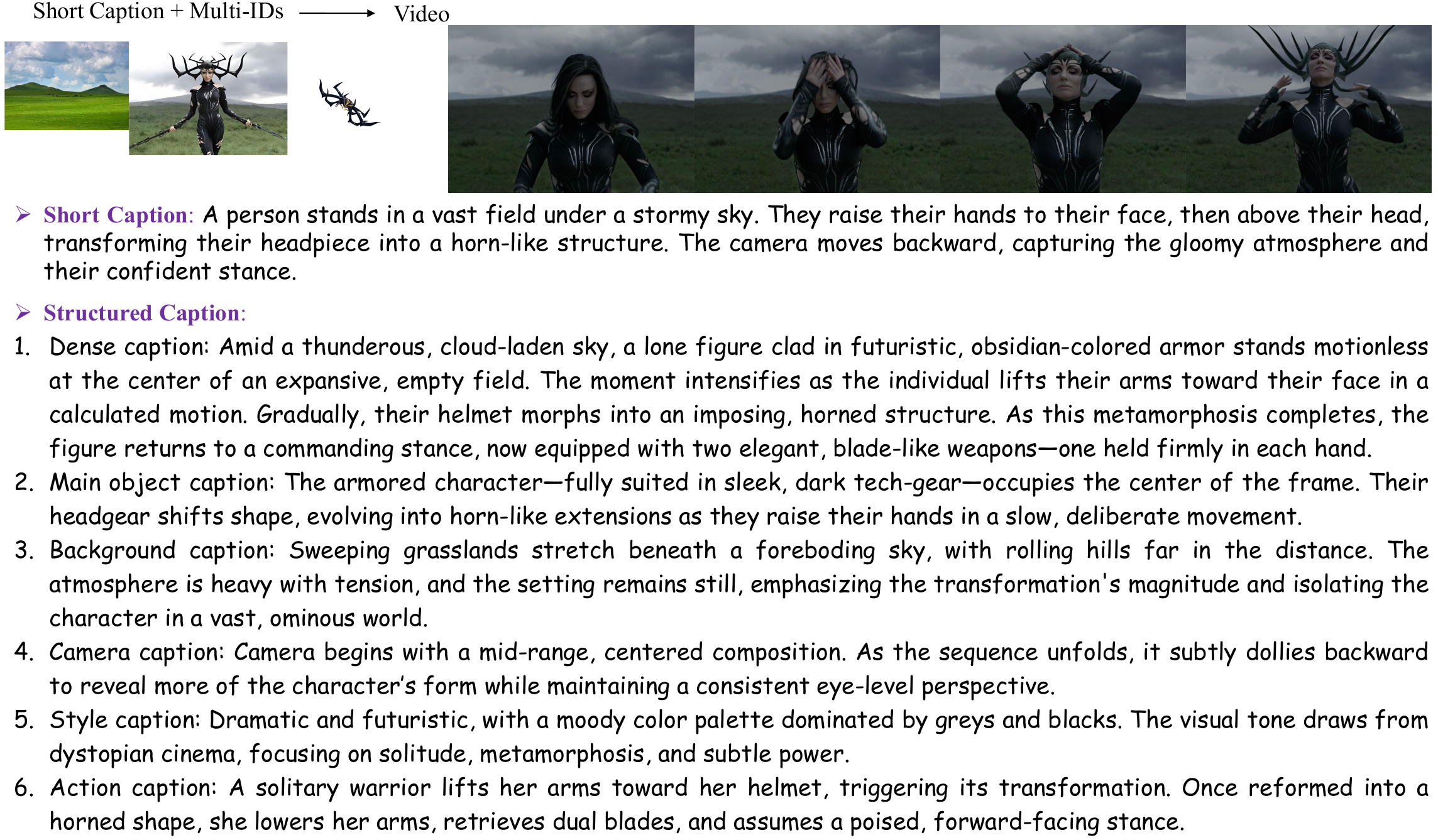}
\vspace{-2mm}
\caption{
    Illustrations of constructed short and structured captions under the multiIDs-to-video generation. 
}
\vspace{-3mm}
\label{fig:caption-case-2}
\end{figure*}

\begin{figure*}[!t]
\centering
\includegraphics[width=0.95\textwidth]{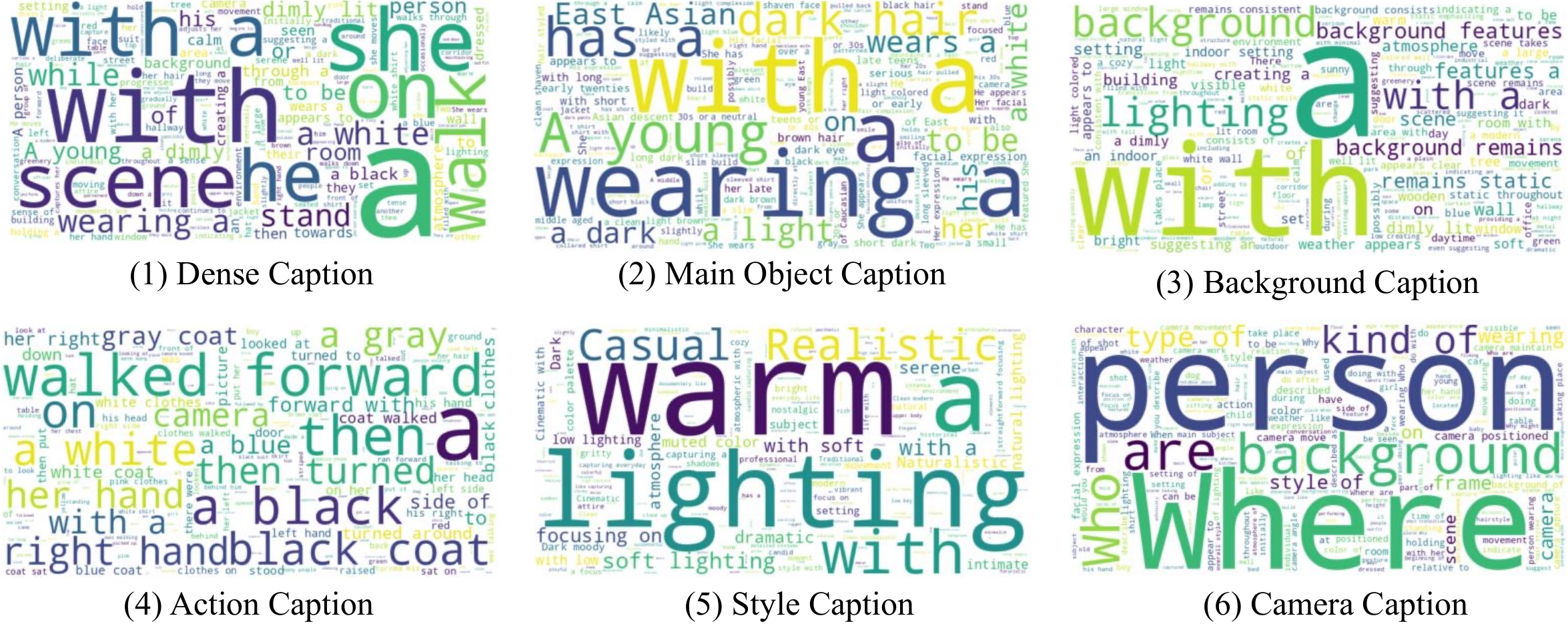}
\vspace{-2mm}
\caption{
    Word cloud of different structured captions in \texttt{Any2CapIns} dataset, showing the diversity.
}
\vspace{-3mm}
\label{fig:wordcloud-cap}
\end{figure*}

\begin{table*}[!th]
\centering
\fontsize{8}{9}\selectfont
\setlength{\tabcolsep}{0.5mm}
\begin{tabular}{lp{16cm}}
\toprule
\textbf{Multi-IDs} & Here is the scenario: We have an MLLM model that supports a text image-conditioned intrinsic-video-caption generation task. The system input consists of:

\quad 1. A reference image composed of 2-3 horizontally stitched images (provided by the user), each stitched image containing one or several target objects for reference); and

\quad 2. A concise textual prompt (referred to as text B, the user's instruction).

The model's output is a detailed descriptive caption (**text A**) that thoroughly describes the video corresponding to the user's input prompt (**text B**) in great detail.
Your task is to perform a reverse engineering. Based on the given reference image (the target objects) and the detailed target video caption (text A), you need to generate a **reasonable and concise user prompt (text B)** through your understanding, analysis, and imagination.
You must adhere to the following rules:

\quad 1. Text A is a dense caption of a video, including all the key objects, their attributes, relationships, background, camera movements, style, and more. Carefully analyze this caption for all relevant details.
    
\quad 2. Analyze the provided reference images in detail to identify the differences or missing details compared to the target video description. These may include environment details, the interaction between objects, the progression of actions, camera movements, style, or any elements not covered by the reference image. Based on these analyses, generate the user's instructions.

\quad 3. The user's prompt must include the following aspects: first, an overall description of where the target objects are and what they are doing, along with the temporal progression of their actions. Then, it should describe the background, style, and camera movements.

\quad 4. If the target video introduces new objects not present in the reference images, the user's prompt should describe the attributes of the new target objects and their interactions with the other target objects.

\quad 5. If the video's style differs from the reference, briefly describe the style in a few words.

\quad 6. When the background needs to be described, include details about people, settings, and styles present in the background.

\quad 7. Avoid repeating information that can be inferred from the reference images, and eliminate redundant descriptions in the user prompt.

\quad 8. The user prompt (text B) must be written in simple wording, maintaining a concise style with short sentences.

\quad 9. The user's instructions should vary in expression;  For example, prompts do not always need to start with the main subject. They can begin with environmental details, camera movements, or other contextual aspects.

Here are three examples representing the desired pattern:

====================================================================================================

\quad \textcolor{blue}{[In-context Examples]}

====================================================================================================

\quad \textcolor{blue}{[Input]}
 \\
\bottomrule
\end{tabular}
\vspace{-2mm}
\caption{
Demonstration of the prompt used for GPT-4V to generate the short prompt when the input condition is the multi-IDs.
}
\label{tab:prompt-multiid}
\end{table*}

\begin{table*}[!th]
\centering
\fontsize{8}{9}\selectfont
\setlength{\tabcolsep}{0.5mm}
\begin{tabular}{lp{16cm}}
\toprule
\textbf{Depth} & Here is the scenario: We have an MLLM model that supports a text \& image-conditioned intrinsic-video-caption generation task. The system input consists of:

\quad 1. A reference image composed of 3-5 horizontally stitched depth maps in temporal sequence (provided by the user, each map containing depth information for reference); and

\quad 2. A concise textual prompt (referred to as text B, the user's instruction).

The model's output is a detailed descriptive caption (text A) that thoroughly describes the video corresponding to the user's input prompt (text B) in great detail.
Now, I need you to perform a reverse engineering task. Based on the given reference image (the depths) and the detailed target video caption (text A), you must generate a reasonable and concise user prompt (text B) through your understanding, analysis, and imagination.
To ensure accurate and effective outputs, follow these rules strictly:

\quad 1. Text A is a dense caption of a video, including all the key objects, their attributes, relationships, background, camera movements, style, and more. Carefully analyze this caption to extract the necessary details.

\quad 2. Since the depth information already provides the necessary geometric outlines and layout details. Do not repeat this information in the user prompt. Instead, focus on the aspects not covered by the depth maps.

\quad 3. The user's instruction should highlight details not included in the depth map, such as environmental details, the appearance of the subjects, interactions between subjects, the progression of actions, relationships between the subjects and the environment, camera movements, and overall style.

\quad 4. For dense depth maps (more than 5 maps), assume the maps provide the camera movements and actions between objects, focusing on describing the appearance of the subjects and environment, the atmosphere, and subtle interactions between subjects and their environment.

\quad 5. For sparse depth maps (5 maps or fewer), assume the maps only provide scene outlines. Emphasize details about the subjects' appearance, environment, interactions between subjects, relationships between subjects and the environment, and camera movements.

\quad 6. The user prompt (text B) must be written in simple wording, maintaining a concise style with short sentences, with a total word count not exceeding 100.

\quad 7. Your output should be a continuous series of sentences, not a list or bullet points.

\quad 8. The user's instructions should vary in expression; they don't always need to begin with a description of the main subject. They could also start with environmental details or camera movements.

Here are three examples representing the desired pattern:

====================================================================================================

\quad \textcolor{blue}{[In-context Examples]}

====================================================================================================

\quad \textcolor{blue}{[Input]}
\\
\bottomrule
\end{tabular}
\vspace{-2mm}
\caption{
Demonstration of the prompt used for GPT-4V to generate the short prompt when the input condition is the depth.
}
\label{tab:prompt-depth}
\end{table*}

\begin{table*}[!th]
\centering
\fontsize{8}{9}\selectfont
\setlength{\tabcolsep}{0.5mm}
\begin{tabular}{lp{16cm}}
\toprule
\textbf{Aspect} & \textbf{Q\textcolor{answerblue}{A} Pairs} \\
\midrule
Main Object & 
What is the young woman adjusting as she walks down the corridor? \quad \textcolor{answerblue}{Her wide-brimmed hat.} 

What color is the young woman's T-shirt?  \quad \textcolor{answerblue}{Light blue.} 

How does the young woman feel as she walks down the corridor? \quad \textcolor{answerblue}{Happy and carefree.} 

What is the young woman wearing? \quad \textcolor{answerblue}{Light blue t-shirt with pink lettering, blue jeans, and a wide-brimmed hat.}

What is the young woman's hair length? \quad \textcolor{answerblue}{Long.}

What is the position of the young woman in the frame? \quad \textcolor{answerblue}{In the center of the frame.}

What is the main object in the video? \quad \textcolor{answerblue}{A large shark.}

What is the color of the underwater scene?\quad \textcolor{answerblue}{Blue.}

What are the two scientists wearing? \quad \textcolor{answerblue}{White lab coats and gloves.}

What is the first scientist using? \quad \textcolor{answerblue}{A microscope.}
\\

\hline

Background  &  Where is the young woman walking?  \quad \textcolor{answerblue}{Down a corridor.} 

What time of day does the scene appear to be set?  \quad \textcolor{answerblue}{Daytime.}

What can be seen in the background of the corridor?   \quad \textcolor{answerblue}{Beige walls and large windows.}

What is the weather like in the video?    \quad \textcolor{answerblue}{Clear.}

Where is the shark located?    \quad \textcolor{answerblue}{On the ocean floor.}

What surrounds the shark in the video? \quad \textcolor{answerblue}{Smaller fish.}

Where is the laboratory setting? \quad \textcolor{answerblue}{In a brightly lit environment with shelves filled with bottles.}

What detail does the background highlight?\quad \textcolor{answerblue}{The scientific setting with static emphasis.}
\\

\hline

Camera & How does the camera follow the young woman? \quad \textcolor{answerblue}{Moving backward} 

What is the camera's height relative to the person?
\quad \textcolor{answerblue}{Roughly the same height as the person.}

What shot type does the camera maintain? \quad \textcolor{answerblue}{Medium close-up shot of the upper body.}

How does the camera position itself to capture the subject? \quad \textcolor{answerblue}{At a higher angle, shooting downward.}

How does the camera capture the environment?  \quad \textcolor{answerblue}{From a medium distance.}

How is the camera positioned?  \quad \textcolor{answerblue}{At approximately the same eye level as the subjects, maintaining a close-up shot.}

How does the camera move in the video?\quad \textcolor{answerblue}{It pans to the right.}
\\

\hline

Style &  

What is the style of the video? \quad \textcolor{answerblue}{Casual and candid.}

What kind of design does the corridor have? \quad \textcolor{answerblue}{Modern and clean design.}

What style does the video portray?  \quad \textcolor{answerblue}{Naturalistic style with clear, vivid visuals.}

What does the video style emphasize?   \quad \textcolor{answerblue}{Clinical, high-tech, and scientific precision.}

What is the color theme of the lighting?    \quad \textcolor{answerblue}{Bright and cool.}

What kind of atmosphere does the laboratory have?    \quad \textcolor{answerblue}{Professional and scientific.}

\\
\hline

Action &  What does the young woman do with both hands occasionally?    \quad \textcolor{answerblue}{Adjusts her hat.} 

What is the young woman doing as she moves?    \quad \textcolor{answerblue}{Walking forward with her hands on her hat.}

What is the main action of the shark in the video?    \quad \textcolor{answerblue}{Lying motionless.}

What is the movement of the fish like? \quad \textcolor{answerblue}{Calm and occasionally darting.}

What is the movement of the first scientist at the beginning?\quad \textcolor{answerblue}{Examines a microscope.}

What task is the second scientist engaged in? \quad \textcolor{answerblue}{Handling a pipette and a beaker filled with green liquid.}

How does the second scientist transfer the liquid? \quad \textcolor{answerblue}{Carefully using a pipette into the beaker.}

Are there any noticeable movements in the background?\quad \textcolor{answerblue}{Occasional small particles floating.}
\\

\bottomrule
\end{tabular}
\vspace{-2mm}
\caption{
Demonstration of generated question-answer pairs utilized in IR\textsc{Score} calculation.
}
\label{tab:qa-pairs}
\end{table*}

\begin{figure}[!h]
    \centering
    \includegraphics[width=0.99\linewidth]{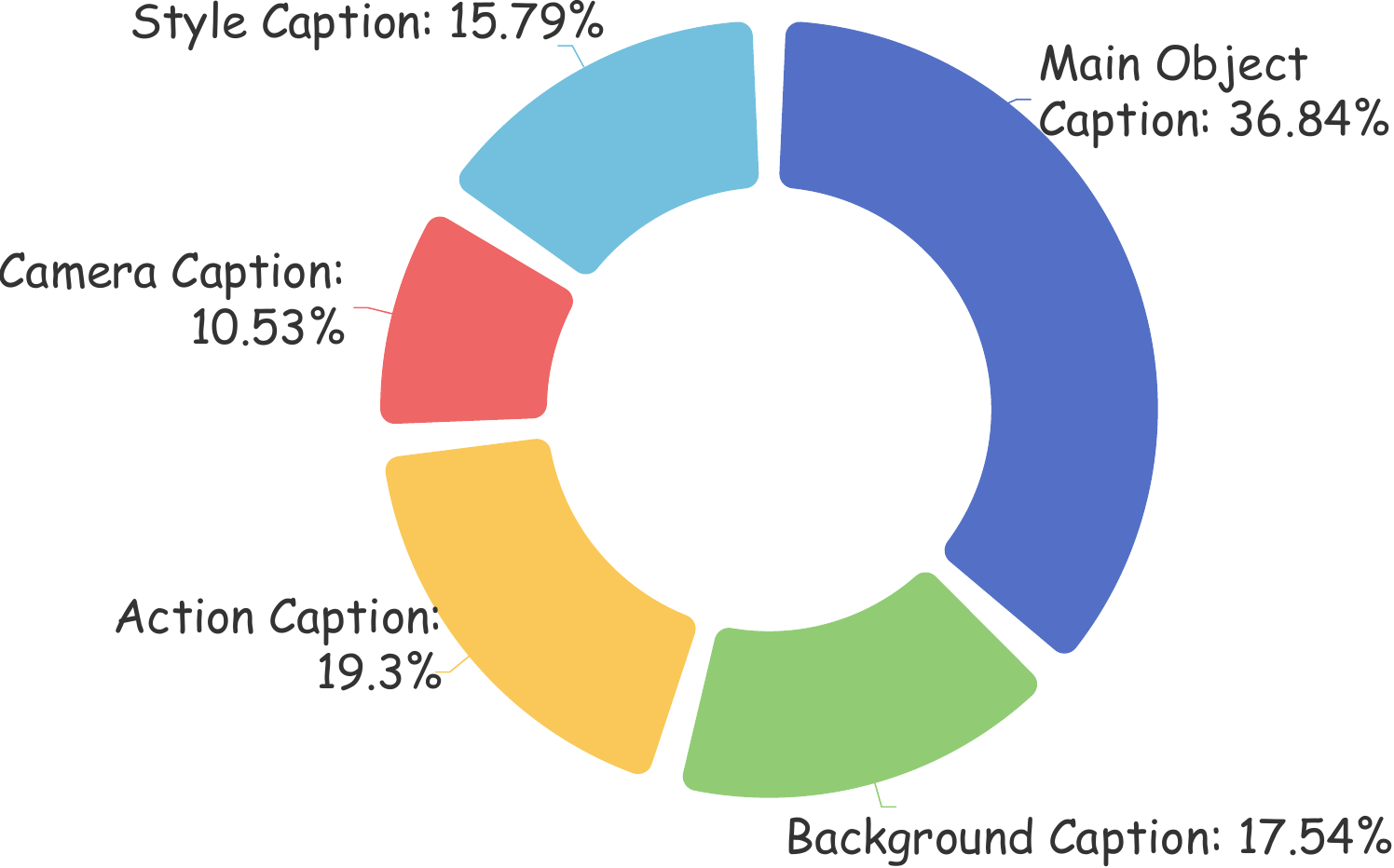}
    \caption{QA pairs proportion in structured captions.}
    \label{fig:qa_dirtribution}
    \vspace{-4mm}
\end{figure}

\section{Extended Dataset Details}
\label{app:dataset_details}

\paragraph{Visualization of Short and Structured Caption.}

Here, we visualize the short and structured caption in Fig. \ref{fig:caption-case-1}, \ref{fig:caption-case-2}.
Notably, the structured caption captures the video content in greater detail and across multiple aspects. 
In contrast, the short caption deliberately omits any information already supplied by a non-text condition—for example, camera movement is excluded in Fig. \ref{fig:caption-case-1}'s short caption but included in Fig. \ref{fig:caption-case-2} because it is not specified by the provided multiple-identity images.
Moreover, we visualize the word distribution of the structured captions in \texttt{Any2CapIns} in Fig. \ref{fig:wordcloud-cap}.

\paragraph{Prompt Visualization for Short Caption Construction.}

In Tab. \ref{tab:prompt-multiid} and \ref{tab:prompt-depth}, we show the system prompts used by GPT-4V to generate short captions. 
The prompt explicitly instructs GPT-4V to consider the given conditions comprehensively and produce short prompts that focus on information not covered by the non-textual inputs. 
For instance, when multiple identities are specified, the short prompt should avoid repeating their appearance attributes and instead highlight interactions among the identities. 
Conversely, when depth is the input condition, the short prompt should include more detailed appearance-related information.

\section{More Statistics Information of IR\textsc{Score}}
\label{app:eval_suite}

We generate a total of 15,378 question-answer (QA) pairs, averaging 19.2 pairs per structured caption. Fig. \ref{fig:qa_dirtribution} presents the distribution of constructed questions across different aspects in the structured caption, and Tab. \ref{tab:qa-pairs} shows representative QA pairs for each aspect. 
Notably, questions under the \emph{main object} category emphasize fine-grained details such as clothing color or hairstyles, while \emph{action} questions focus on object interactions and movements. 
This level of specificity allows us to rigorously assess whether the generated captions are both complete and precise.

\section{Detailed Setups}
\label{app:setups}

\subsection{Detailed Testing Dataset}

Here, we present the statistics of the test in Tab. \ref{tab:test_statistics}, which covers four types of single conditions (e.g., \textit{Depth}, \textit{Camera}, \textit{Identities}, and \textit{Human pose}), and four types of compositional conditions(e.g., \textit{Camera+Identities}, \textit{Camera+Depth}, \textit{Identities+Depth} and \textit{Camera+Identities+Depth}).
Each category contains 200 instances.

\begin{table}[!h]
\centering
\fontsize{8}{9}\selectfont
\setlength{\tabcolsep}{0.2mm}
\begin{tabular}{lcccc}
\toprule
    \multirow{2}{*}{\textbf{Type}} & \multirow{2}{*}{\textbf{\#Inst.}} & \multirow{2}{*}{\textbf{\#Condi.}}  & \textbf{Short Cap. }  & \textbf{\#Structured Cap.} \\
    & & & \textbf{\#Avg. Len.} & \textbf{\#Avg. Len.} \\
\midrule
   Identities  & 200 & 350 & 65.28  & 284.97 \\
   Camera  & 200 & 200 & 50.25  & 208.01 \\
   Depth  & 200 & 200 & 54.22  & 225.09 \\
   Human Pose  & 200 & 200 & 58.38  & 259.03 \\
   Camera+Identities  & 200 & 622 & 53.41 & 209.17 \\
   Camera+Depth  & 200 & 400 & 51.43  & 208.81\\
   Identities+Depth  & 200 & 555 & 53.14 & 286.83  \\
   Camera+Identities+Depth  & 200 & 756 & 58.35 & 289.21  \\
\bottomrule
\end{tabular}
\caption{
Statistics of the constructed test datasets. \textbf{\#Inst.} denotes the number of instances, and \textbf{\#Condi.} indicates the number of unique conditions. \textbf{Short Cap. \#Avg. Len} represents the average caption length of short captions, and \textbf{Structured Cap. \#Avg. Len.} represents the average caption length of structured captions.
}
\label{tab:test_statistics}
\end{table}

\begin{table*}[!h]
\centering
\fontsize{8}{9}\selectfont
\setlength{\tabcolsep}{1.5mm}
\begin{tabular}{l cc cccc}
\toprule
\multirow{2}{*}{\textbf{Configuration}} & \multicolumn{2}{c}{\textbf{Stage-1: Alignment Learning}} & \multicolumn{4}{c}{\textbf{Stage-2: Condition-Interpreting Learning}}  \\
\cmidrule(r){2-3}\cmidrule(r){4-7}
& Camera & Motion & Identities & Human pose & Camera & Depth \\    
\midrule
Optimizer  & AdamW & AdamW & AdamW  & AdamW & AdamW & AdamW \\
Precision & bfloat16  & bfloat16  & bfloat16  & bfloat16  & bfloat16 & bfloat16\\
Learning Rate & 5e\-5 & 5e\-5 & 5e\-5 & 5e\-5 & 5e\-5 & 1e\-5\\
Weight Decay & 0.01 & 0.01 & 0.01 & 0.01 & 0.01 & 0.01 \\
Joint Train Ratio & 0.0 & 0.0 & 0.0 & 0.4 & 0.6 & 0.8\\
Betas & (0.9, 0.99) & (0.9, 0.99) & (0.9, 0.99) & (0.9, 0.99) & (0.9, 0.99) \\
Dropout Possibility & 0.0 & 0.0 & 0.4 & 0.6 & 0.6 & 0.6 \\
Dropout (Short Cap.) & 0.0 &  0.0 & 0.6 & 0.6 & 0.6 & 0.6 \\
Batch Size Per GPU & 4 & 4 & 4 & 4 & 4 & 4 \\
Training Data & \makecell{Camera Movement Description \\ Dataset (Manual)} & \makecell{Action Description \\ Dataset (Manual)} & MultiIDs & \makecell{Human Pose \\ LLaVA-150K} & \makecell{Camera \\ LLaVA-150K } & \makecell{Depth \\ LLaVA-150K \\ Alpaca-50K}\\
\bottomrule
\end{tabular}
\caption{
Training recipes for \texttt{Any2Caption}.
}
\label{tab:training_receipts}
\end{table*}

\begin{figure*}[!t]
\centering
\includegraphics[width=0.99\textwidth]{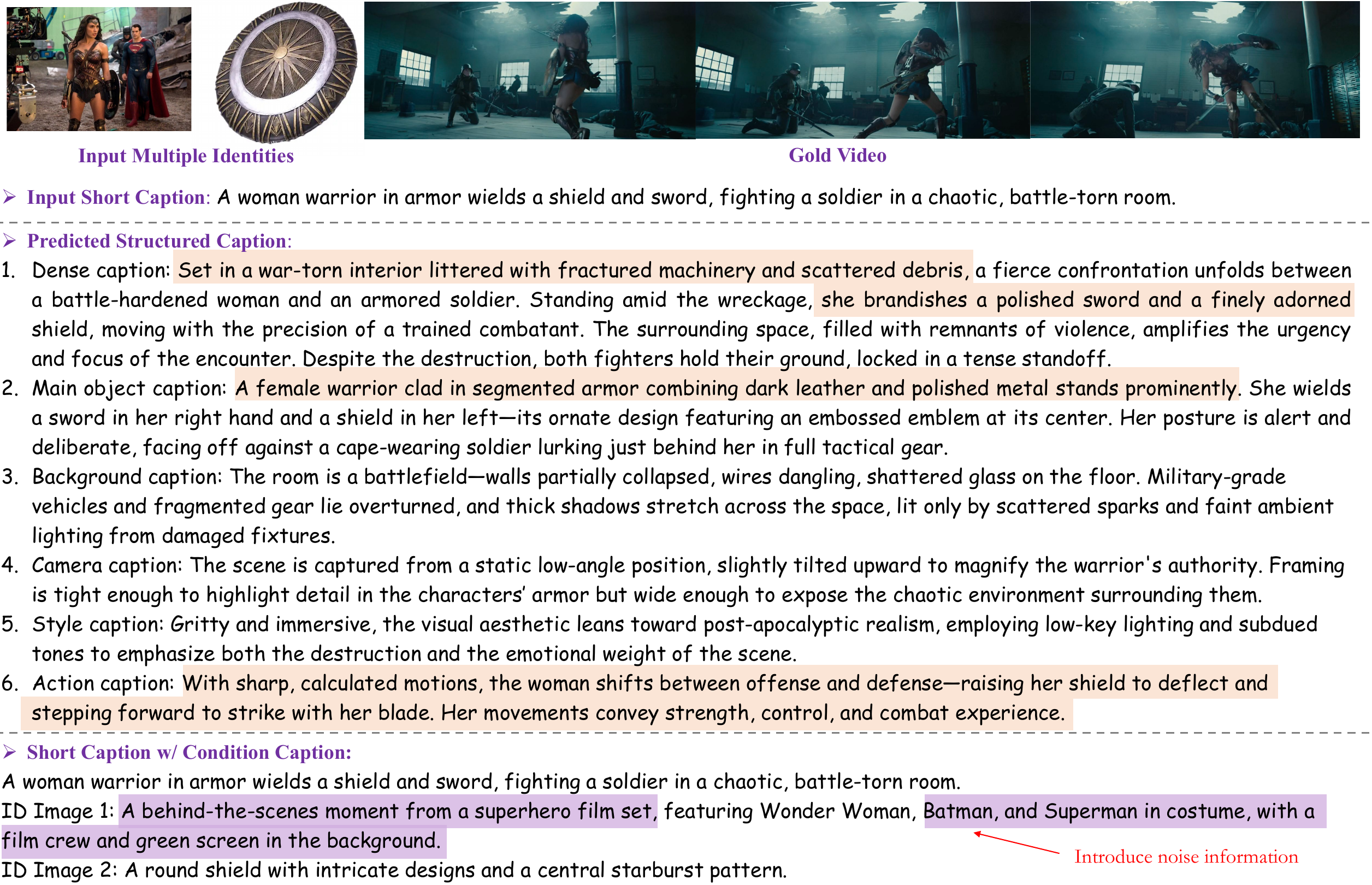}
\vspace{-2mm}
\caption{
    Illustrations of input short caption, predicted structured caption, and short caption combined with condition captions. The condition captions may introduce noisy information due to the focus distraction of the background of input identities, for example, the \textit{Batman}, \textit{Superman}, \textit{film crew}. 
}
\vspace{-3mm}
\label{fig:caption-gen-case-1}
\end{figure*}

\begin{figure*}[!th]
\centering
\includegraphics[width=0.99\textwidth]{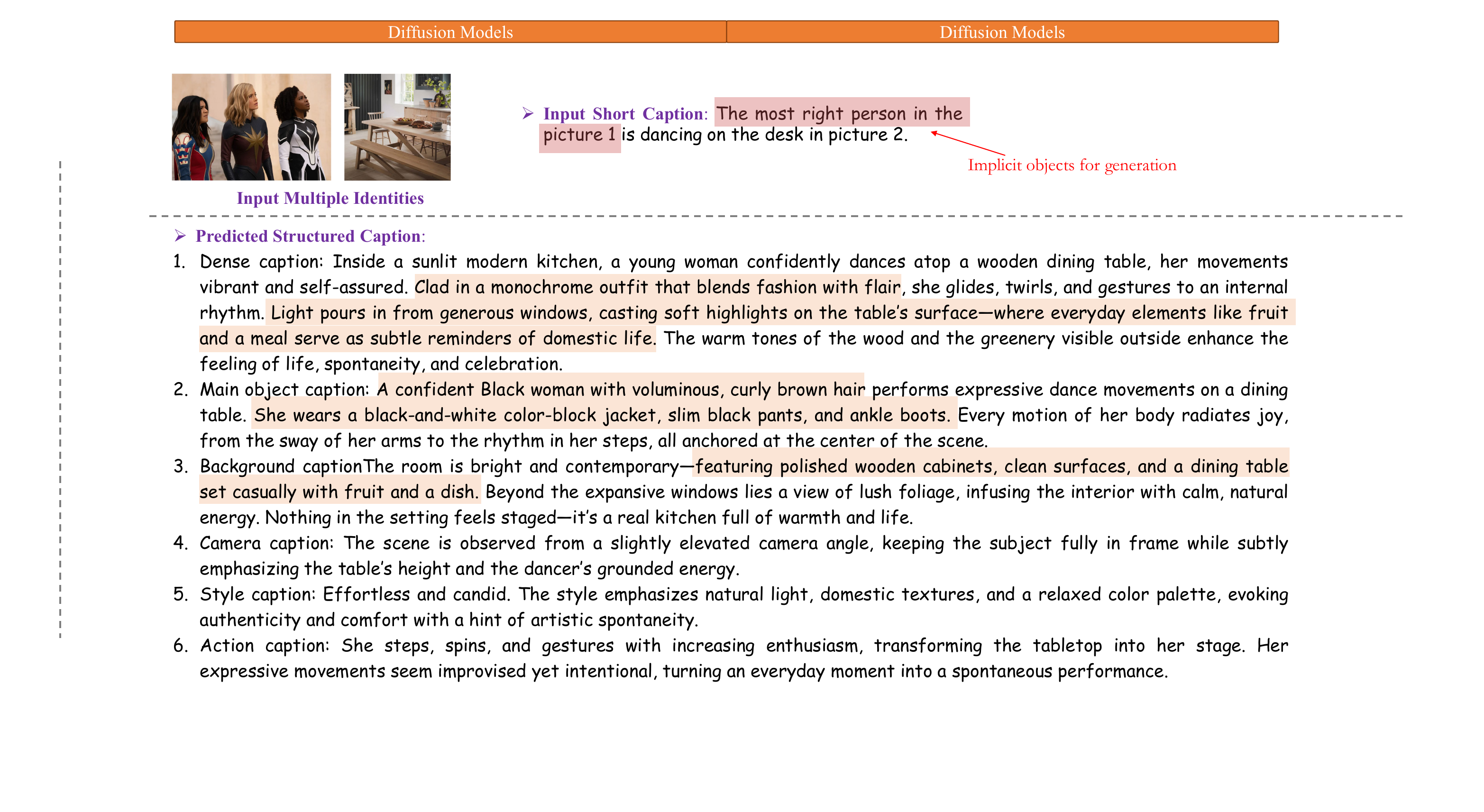}
\vspace{-2mm}
\caption{
    Illustrations of predicted structured captions based on the input multiple identities and the short instruction that expresses the implicit objects for the video generation.
}
\vspace{-3mm}
\label{fig:caption-gen-case-2}
\end{figure*}

\begin{figure*}[!th]
\centering
\includegraphics[width=0.99\textwidth]{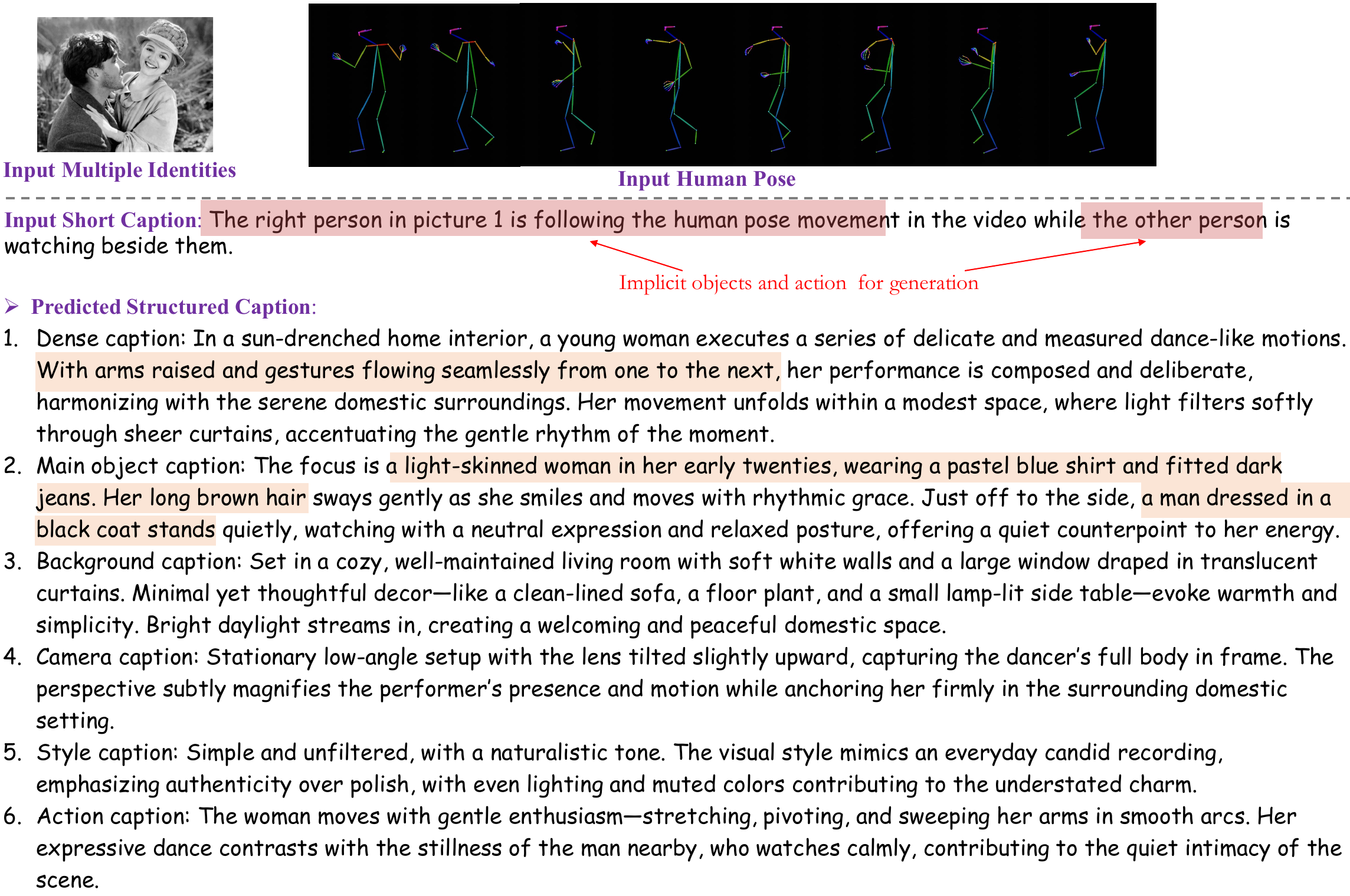}
\vspace{-2mm}
\caption{
    Illustrations of predicted structured captions based on the input multiple identities and the short instruction that expresses the implicit objects and the action for the target video generation.
}
\vspace{-3mm}
\label{fig:caption-gen-case-3}
\end{figure*}

\subsection{Detailed Training Procedure}

We employ a two-stage training process to enhance the alignment and interpretability of multimodal conditions in \texttt{Any2Caption}.

\paragraph{Stage-1: Alignment learning.}

This stage focuses on aligning features extracted by the camera encoder with the LLM feature space. 
To achieve this, we first extract camera movement descriptions (e.g., \textit{fixed}, \textit{backward}, \textit{pan to the right}) from the camera captions in \texttt{Any2CapIns} to construct a camera movement description dataset. We then introduce two specialized tokens, \texttt{<|camera\_start|>} and \texttt{<|camera\_end|>}, at the beginning and end of the camera feature embeddings. During training, only the camera encoder is optimized, while all other parameters in \texttt{Any2Caption} remain frozen.
Similarly, for motion alignment, we construct a motion description dataset by extracting action descriptions (e.g., \textit{walking}, \textit{dancing}, \textit{holding}) from the action captions in \texttt{Any2CapIns}. We then freeze all model parameters except those in the motion encoder to ensure the LLM effectively understands motion-related conditions.

\paragraph{Stage-2: Condition-Interpreting Learning.}

After alignment learning, we initialize \texttt{Any2Caption} with the pre-trained Qwen2-VL, motion encoder, and camera encoder weights. 
We then employ a progressive mixed training strategy, updating only the \texttt{lm\_head} while keeping the multimodal encoders frozen. 
The training follows a sequential order based on condition complexity: identities $\Rightarrow$ human pose $\Rightarrow$ camera $\Rightarrow$ depth. Correspondingly, the integration ratio of additional vision/text instruction datasets is progressively increased, set at 0.0, 0.4, 0.6, and 0.8, ensuring a balanced learning process between condition-specific specialization and generalization.

\subsection{Detailed Implementations}

In Tab. \ref{tab:training_receipts}, we list the detailed hyperparameter settings in two stages.
All the training is conducted on 8$\times$80G GPUs.

\section{Extended Experiment Results and Analyses}
\label{app:experiments}

\subsection{The Capability for Understanding Complex Instruction.}
\label{app:in-depth-analyses}

We further examine \texttt{Any2Caption}'s ability to handle complex user instructions, particularly regarding whether it accurately captures the user’s intended generation targets. From Fig. \ref{fig:caption-gen-case-1}, we observe that the model focuses precisely on the user-specified main objects, such as a ``woman warrior'' or a background ``filled with chaos and destruction''—when producing structured captions. 
In contrast, a short caption combined with condition captions often includes extraneous objects or background details present in the identity images, which distract from the user's intended targets in the final video generation.

Additionally, we assess the model’s performance on instructions containing implicit objects or actions, as shown in Fig. \ref{fig:caption-gen-case-2} and \ref{fig:caption-gen-case-3}. In these examples, the model correctly interprets phrases like ``the most right person'' as ``a young Black woman with long, curly brown hair, wearing a black and white outfit'' and similarly associates implicitly specified objects with the provided conditions, generating structured captions that align with the user's goals.

Lastly, Fig. \ref{fig:video-case-3} compares videos generated using different captions. 
The results indicate that structured captions significantly improve both the smoothness of motion and the overall consistency of the generated videos.

\subsection{More Video Visualization}
\label{app:vis}

\begin{figure*}[!th]
\centering
\includegraphics[width=0.99\textwidth]{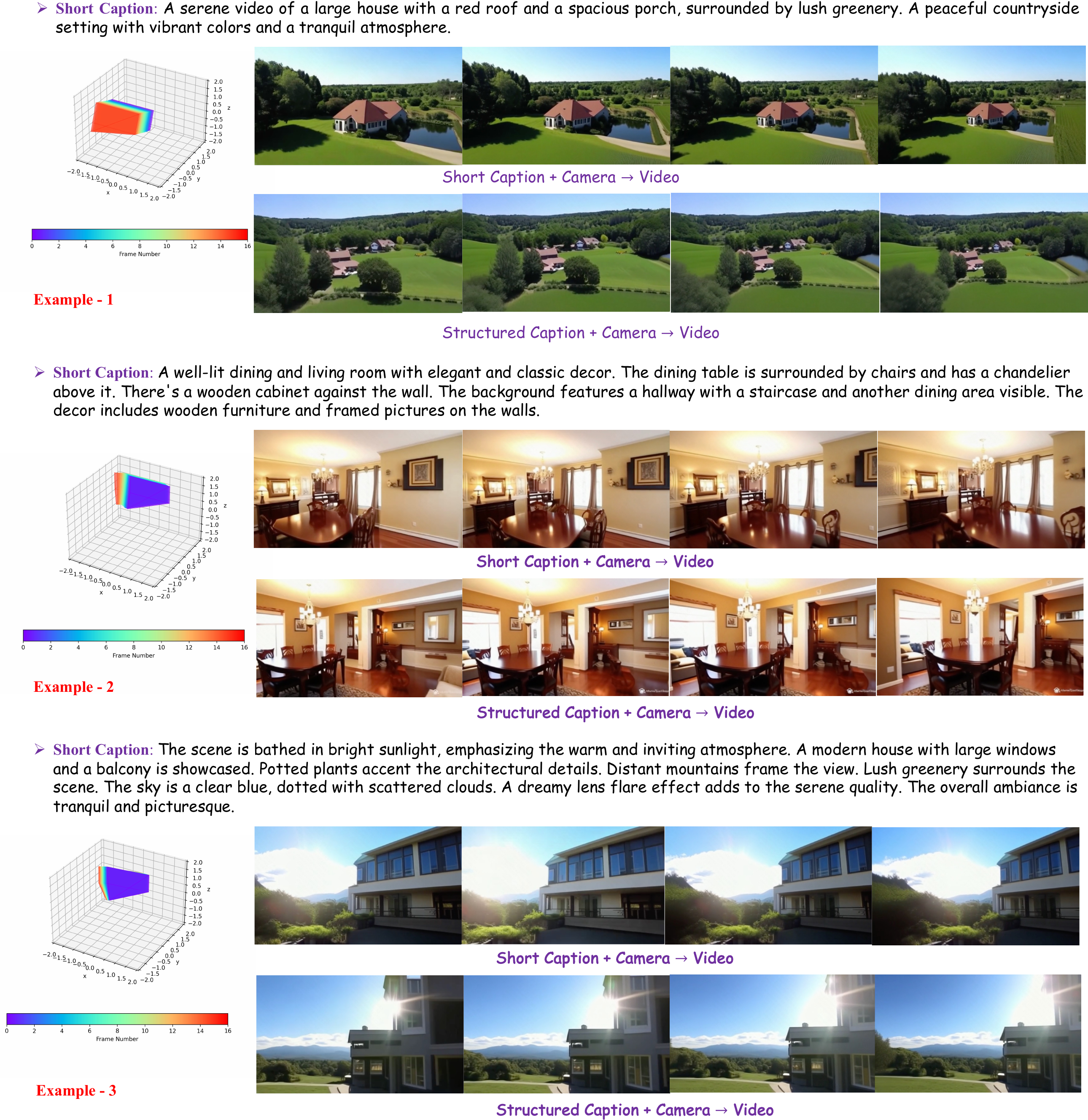}
\vspace{-2mm}
\caption{
    Illustrations of predicted structured captions based on the input multiple identities and the short instruction that expresses the implicit objects and the action for the target video generation.
}
\vspace{-3mm}
\label{fig:video-case-1}
\end{figure*}

\begin{figure*}[!th]
\centering
\includegraphics[width=0.99\textwidth]{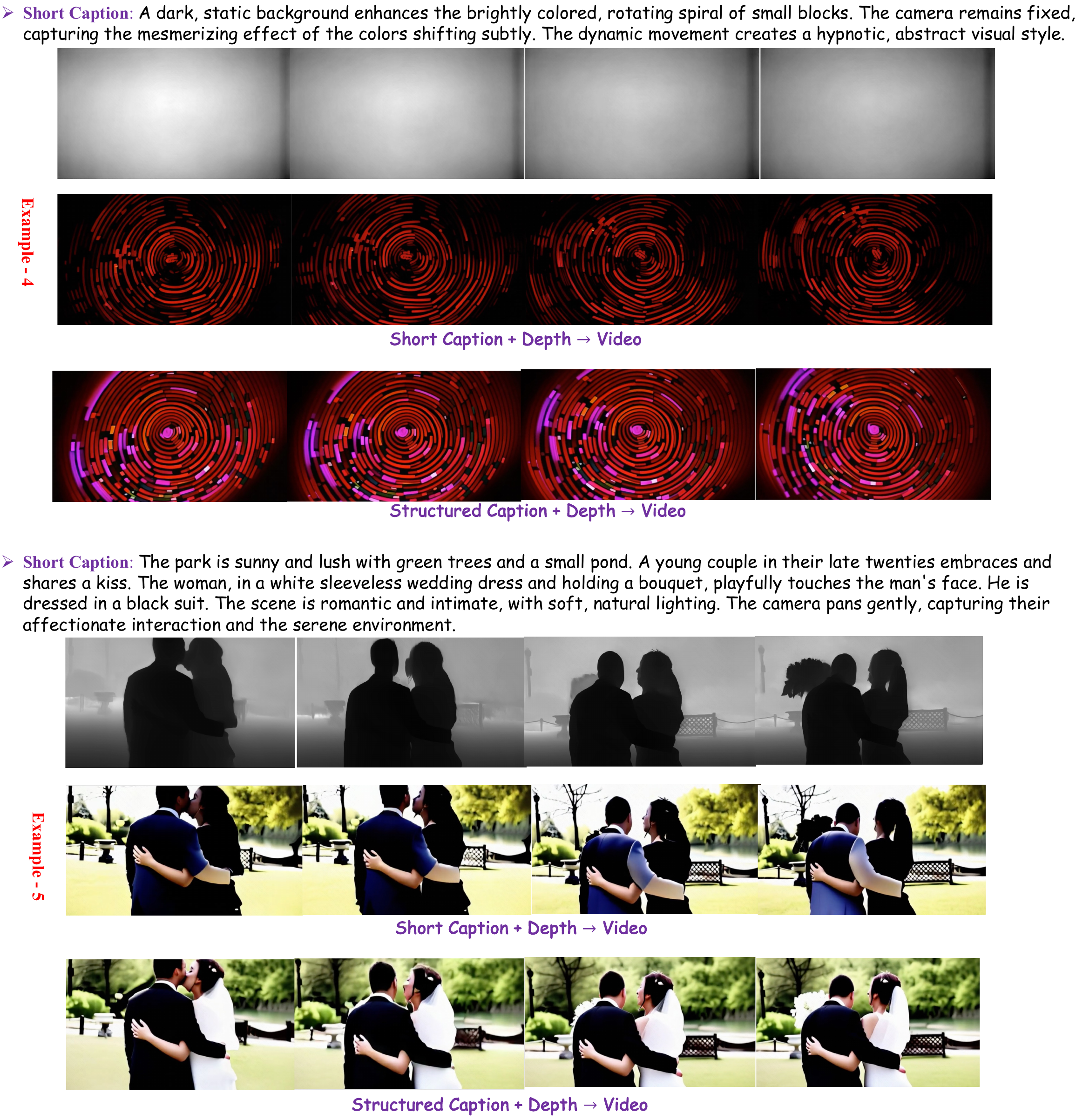}
\vspace{-2mm}
\caption{
    Illustrations of predicted structured captions based on the input multiple identities and the short instruction that expresses the implicit objects and the action for the target video generation.
}
\vspace{-3mm}
\label{fig:video-case-2}
\end{figure*}

Here, we compare the results of different customized models~\cite{ju2025fulldit} after integrating structured captions. Fig. \ref{fig:video-case-1} shows camera-controlled video generation results, and Fig. \ref{fig:video-case-2} illustrates depth-controlled outcomes. We observe that structured captions improve image quality and motion smoothness by providing richer scene details.

\begin{figure*}[!th]
\centering
\includegraphics[width=0.99\textwidth]{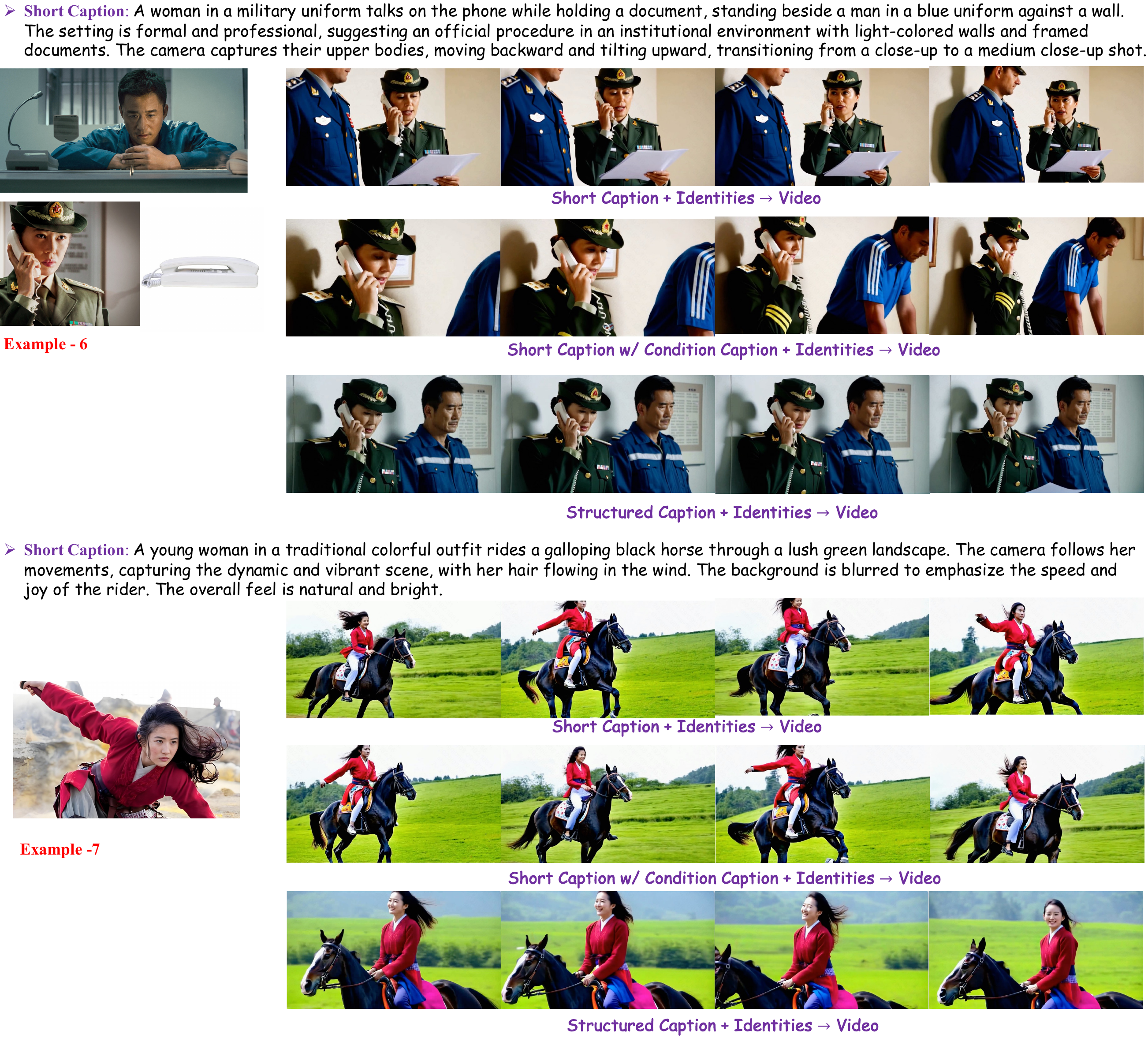}
\vspace{-2mm}
\caption{
    Illustrations of predicted structured captions based on the input multiple identities and the short instruction that expresses the implicit objects and the action for the target video generation.
}
\vspace{-3mm}
\label{fig:video-case-3}
\end{figure*}

\begin{figure*}[!th]
\centering
\includegraphics[width=0.99\textwidth]{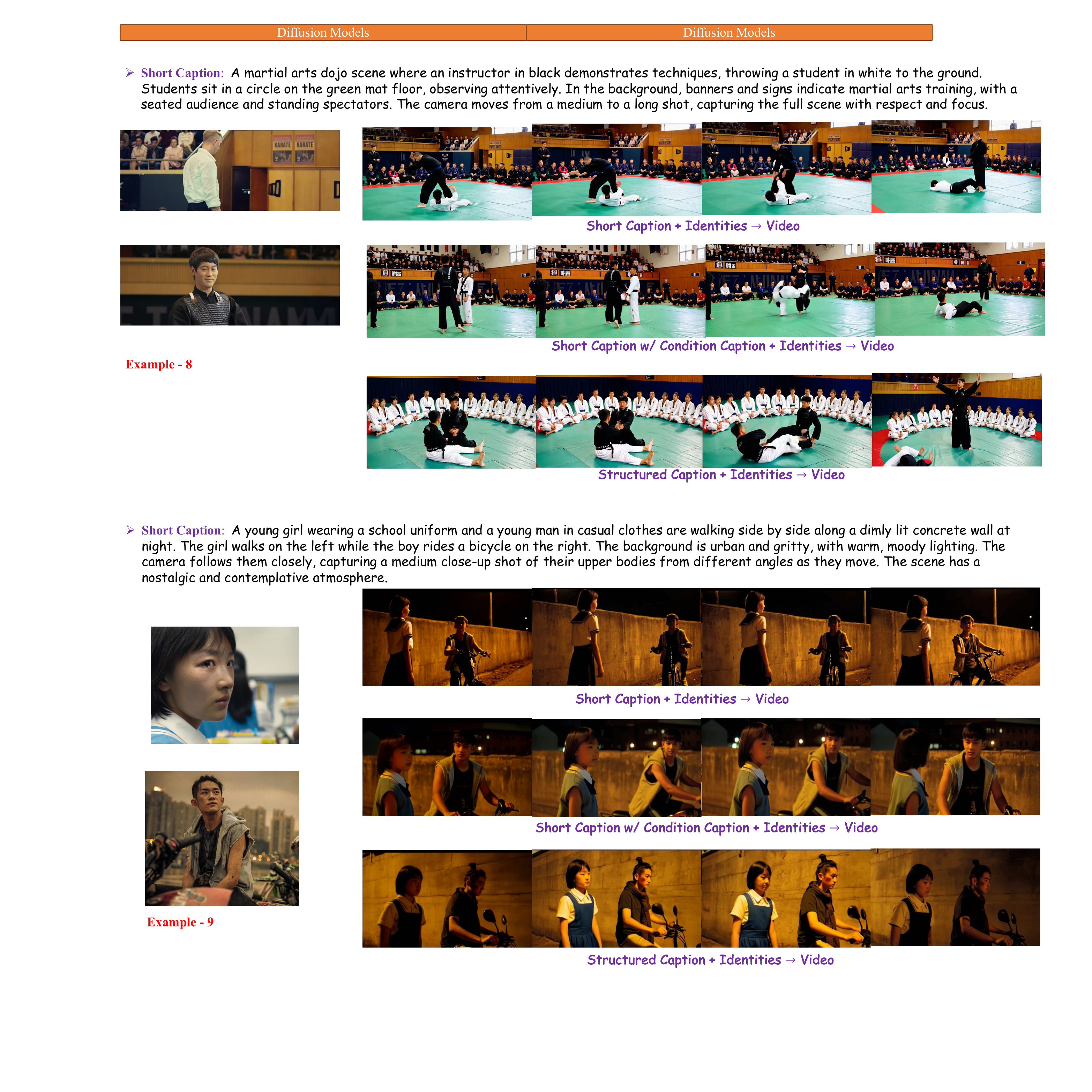}
\vspace{-2mm}
\caption{
    Illustrations of predicted structured captions based on the input multiple identities and the short instruction that expresses the implicit objects and the action for the target video generation.
}
\vspace{-3mm}
\label{fig:video-case-4}
\end{figure*}

\begin{figure*}[!th]
\centering
\includegraphics[width=0.99\textwidth]{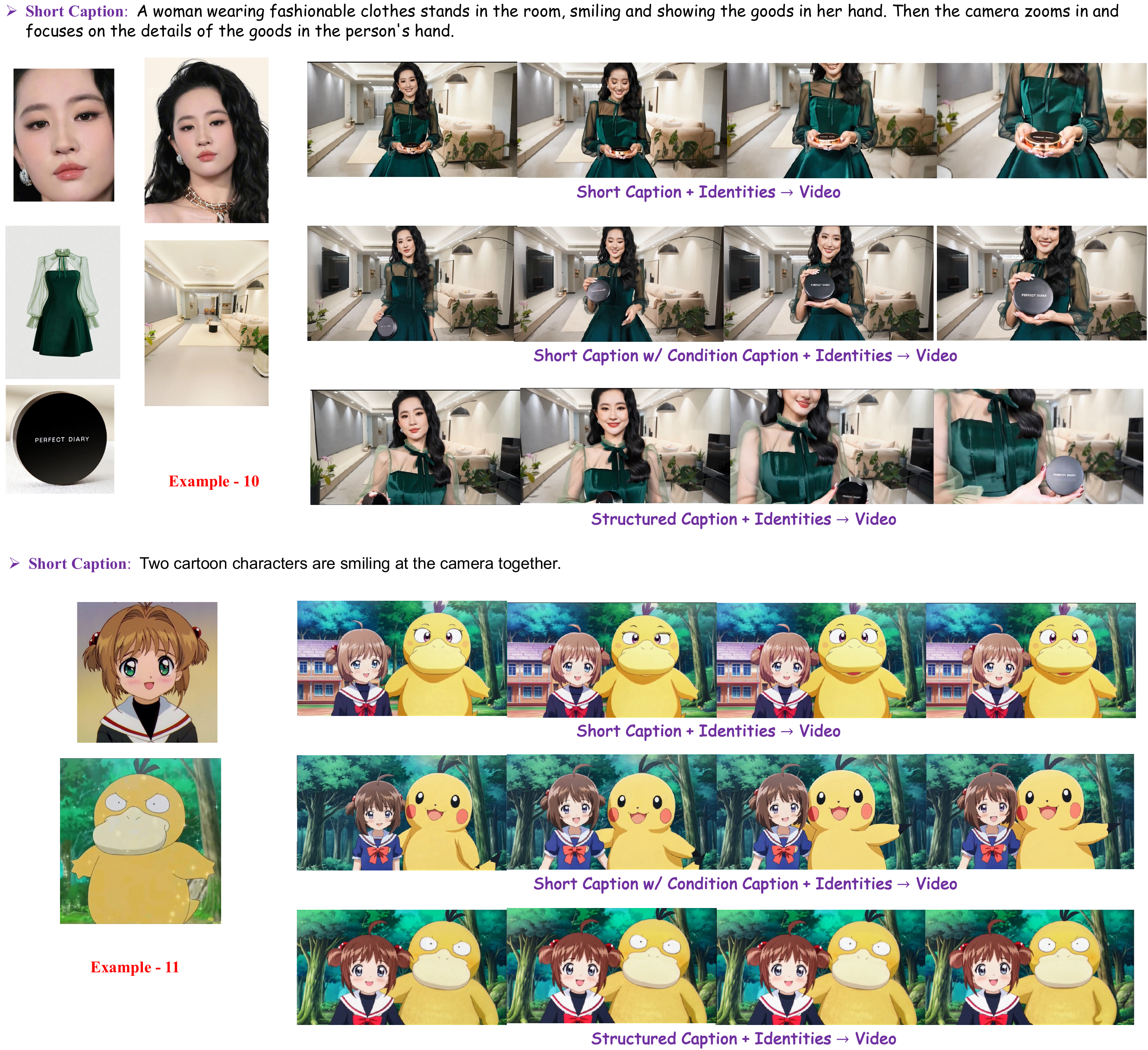}
\vspace{-2mm}
\caption{
    Illustrations of predicted structured captions based on the input multiple identities and the short instruction that expresses the implicit objects and the action for the target video generation.
}
\vspace{-3mm}
\label{fig:video-case-5}
\end{figure*}

For multi-identity-controlled video generation, as depicted in Fig. \ref{fig:video-case-3}, \ref{fig:video-case-4}, and \ref{fig:video-case-5}, structured captions lead to more expressive and realistic representations of the referenced identities, with more accurate colors and smoother actions.

\end{document}

%% file: preamble.tex
\usepackage{multirow}
\usepackage{arydshln}
\usepackage[normalem]{ulem}
\usepackage{color}
\usepackage{graphicx}
\usepackage{amsmath,bm}
\usepackage{amssymb}
\usepackage{array}
\usepackage{comment}
\usepackage{booktabs}
\usepackage{caption}
\usepackage{colortbl} 
\usepackage{multirow}
\usepackage{makecell}
\usepackage{pifont}
\usepackage{tabularray}
\usepackage{paralist}
\usepackage{adjustbox}
\usepackage{graphicx}
\usepackage{rotating}
\usepackage{enumerate}
\usepackage{CJKutf8}
\usepackage{algorithm}
\usepackage{algpseudocode}
\definecolor{lightgreen}{RGB}{196,250,222}
\definecolor{lightblue}{RGB}{211,227,253}
\definecolor{lightgrey}{RGB}{227,229,234}
\definecolor{lightgreyTest}{RGB}{227,229,200}
\definecolor{titlegreen}{RGB}{8,201,185}
\definecolor{pinkBG}{RGB}{255,191,185}
\definecolor{cogvideo}{RGB}{233,113,50}
\definecolor{hunyuan}{RGB}{112,48,160}

\definecolor{questionblue}{RGB}{32,70,178}
\definecolor{answerblue}{RGB}{32,70,178}

\newcommand{\AnyCapLogo}{\xspace\raisebox{-5.2pt}{\includegraphics[width=1.9em]{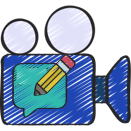}}\xspace}